\documentclass[10pt,twocolumn,letterpaper]{article}

\usepackage{cvpr}              
\usepackage[dvipsnames, table]{xcolor}

\usepackage{graphicx}
\usepackage{amsmath}
\usepackage{amssymb}
\usepackage{booktabs}
\usepackage{mathtools}
\usepackage[accsupp]{axessibility}  
\usepackage{color}
\usepackage{array}
\usepackage{pdflscape}
\usepackage{enumitem}
\usepackage[longtable]{multirow}
\usepackage{longtable}
\usepackage{tabularray}
\usepackage{booktabs}
\usepackage{times}
\usepackage{epsfig}
\usepackage{amsmath}
\usepackage{bbm}
\DeclareMathAlphabet\mathbfcal{OMS}{cmsy}{b}{n}
\usepackage{makecell}
\usepackage{siunitx}
\usepackage{svg}
\usepackage{float}
\usepackage{comment}
\usepackage{lipsum}
\usepackage{stfloats}
\usepackage{multicol}
\usepackage{multirow}
\usepackage{bm}
\usepackage{etoolbox}
\usepackage{icomma}
\usepackage{array}
\usepackage{tabulary}
\usepackage{xcolor}
\usepackage{paralist}
\usepackage{booktabs}
\usepackage{adjustbox}
\usepackage{caption}
\usepackage{subcaption}
\usepackage{arydshln}
\usepackage{rotating}

\usepackage{tcolorbox}  
\usepackage{listings}   
\usepackage{float}      
\usepackage{xcolor}     

\definecolor{codegreen}{rgb}{0,0.6,0}
\definecolor{codegray}{rgb}{0.5,0.5,0.5}
\definecolor{codepurple}{rgb}{0.58,0,0.82}
\definecolor{backcolour}{rgb}{0.95,0.95,0.95}

\definecolor{gray}{rgb}{0.3,0.3,0.3}
\definecolor{blue}{rgb}{0,0.5,1}
\definecolor{mask_red}{rgb}{1,0,0.8}
\definecolor{green}{rgb}{0.2,1,0.2}
\definecolor{rblue}{rgb}{0,0,1}
\definecolor{lightblue}{HTML}{6495ed}
\definecolor{lightred}{HTML}{F19C99}

\definecolor{graytablerow}{gray}{0.6}

\usepackage{pifont}
\newcommand{\cmark}{\ding{51}}%
\usepackage{tabu}
\usepackage[pro]{fontawesome5}

\usepackage{tikz}

\usepackage{enumitem}

\definecolor{cvprblue}{rgb}{0.21,0.49,0.74}
\usepackage[pagebackref,breaklinks,colorlinks]{hyperref}
\hypersetup{colorlinks, citecolor=cvprblue}
\usepackage{tikz}

\def\eg{\emph{e.g}\onedot} 

\def\ie{\emph{i.e}\onedot}

\usepackage[capitalize]{cleveref}
\crefname{section}{Sec.}{Secs.}
\Crefname{section}{Section}{Sections}
\Crefname{table}{Table}{Tables}
\crefname{table}{Tab.}{Tabs.}

\def\paperID{43546} 
\def\confName{CVPR}
\def\confYear{2026}

\title{
More than the Sum: Panorama-Language Models for Adverse Omni-Scenes 
}

\author{Weijia Fan$^{1,3,{\dag}}$ \quad 
Ruiping Liu$^1$ \quad 
Jiale Wei$^1$ \quad 
Yufan Chen$^1$ \quad
Junwei Zheng$^{1,*}$ \quad 
Zichao Zeng$^{1,4}$ \quad \\
Jiaming Zhang$^{2,*}$ \quad 
Qiufu Li$^3$ \quad 
Linlin Shen$^3$ \quad 
Rainer Stiefelhagen$^1$ \quad \\
\normalsize
$^1$Karlsruhe Institute of Technology
\normalsize \quad
$^2$Hunan University
\normalsize \quad
$^3$Shenzhen University
\normalsize \quad
$^4$UCL
}

\begin{document}
\twocolumn[{%
\renewcommand\twocolumn[1][]{#1}%
\maketitle
\begin{center}
    \centering
    \captionsetup{type=figure}
    \centering
    \begin{subfigure}[t]{0.33\textwidth}
        \centering
        \includegraphics[width=\textwidth]{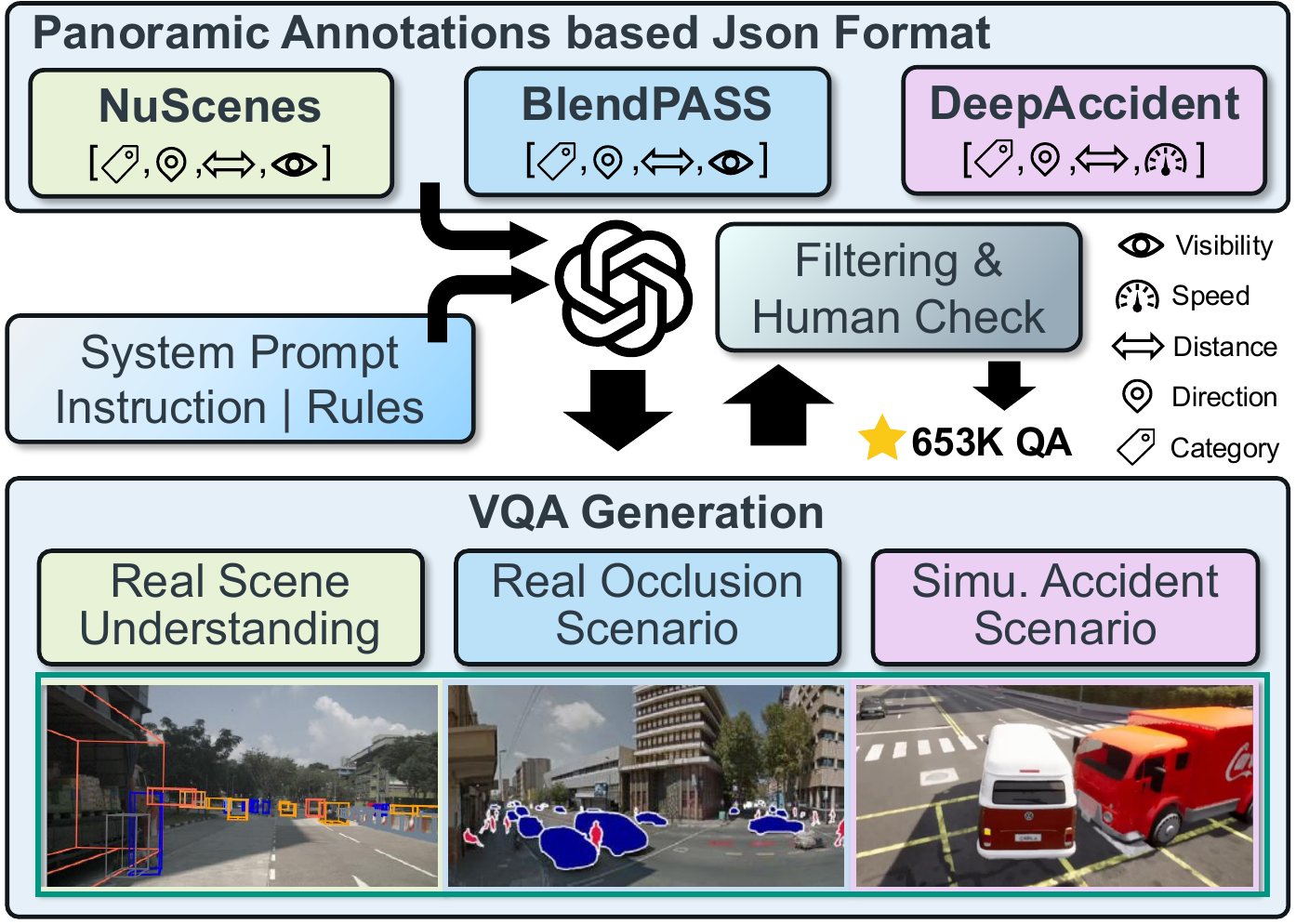}
        \caption{Our PanoVQA generating pipeline.}\label{fig:fig1_a}
    \end{subfigure}
    \begin{subfigure}[t]{0.2672\textwidth}
        \centering
        \includegraphics[width=0.98\textwidth]{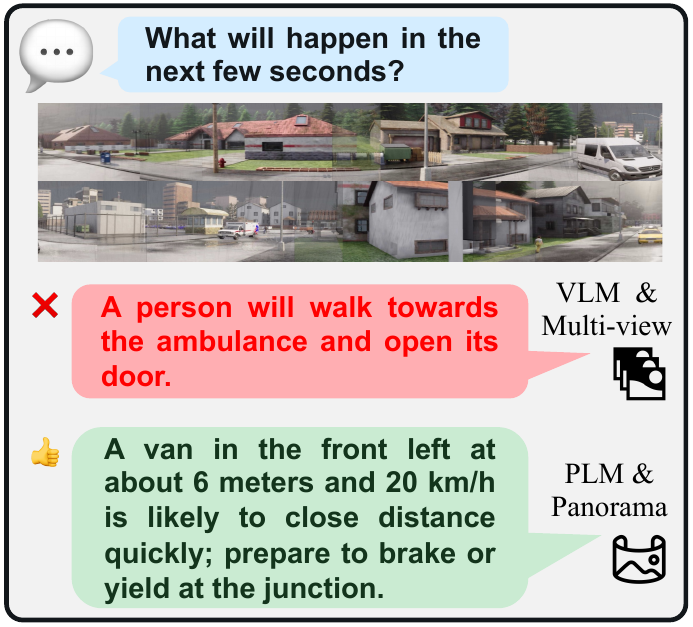}
        \caption{Comparison between VLM and PLM.}\label{fig:fig1_b}
    \end{subfigure}
     \begin{subfigure}[t]{0.40\textwidth}
        \centering
        \includegraphics[width=1.\textwidth]{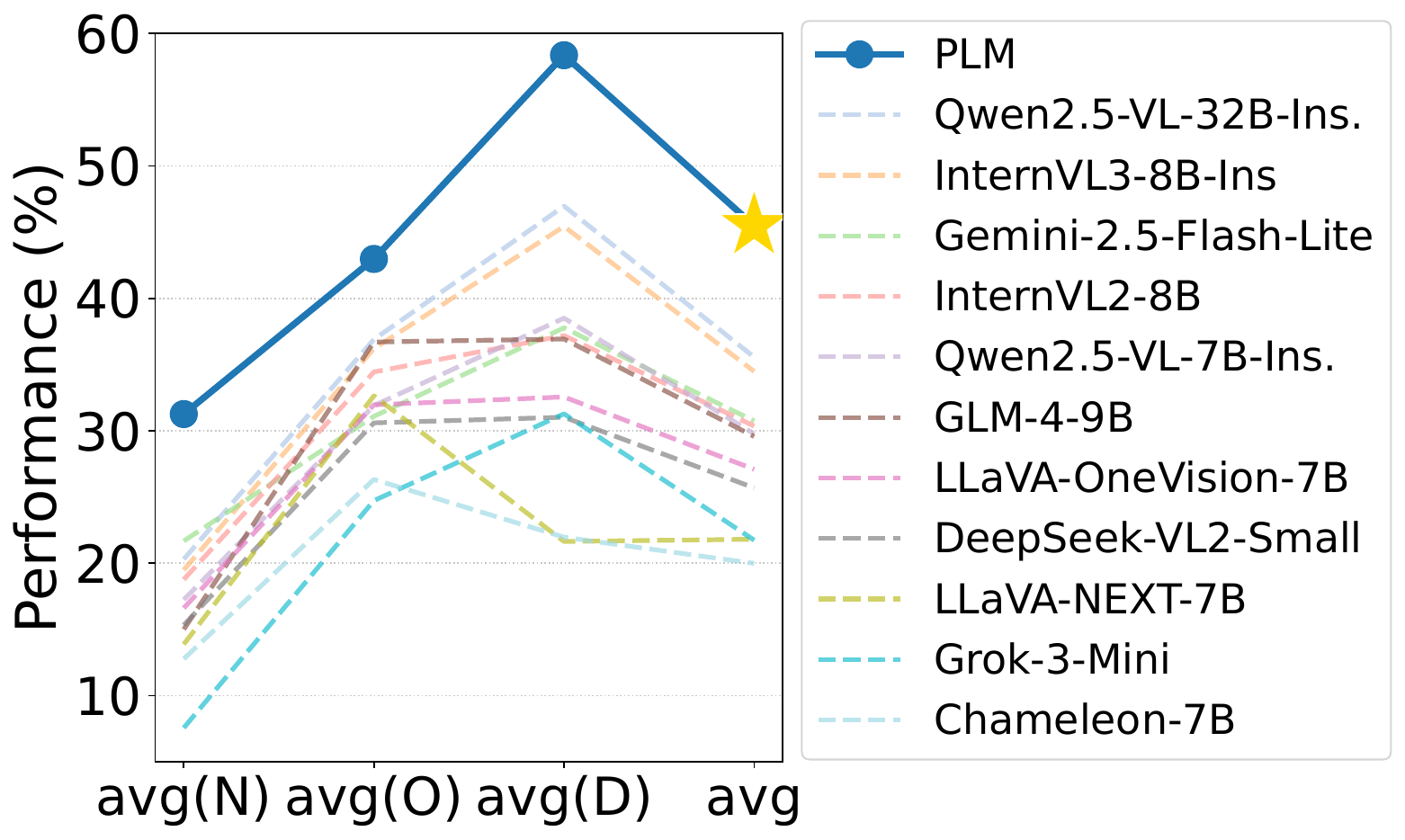}
        \caption{Performance of VLMs and PLM on benchmark.} \label{fig:fig1_c}
    \end{subfigure}
    \setcounter{figure}{0} 
    \captionof{figure}{
    Overview of \textbf{Panorama-Language Modeling (PLM)}. (a) To enable PLM, we create the first PanoVQA dataset with 653K QA pairs, including {normal (N), occluded (O), accidental (D)} driving scenarios. (b) Compared to narrow-FoV multi-view VLMs, PLM with $360^\circ$ spatial semantic consistency can identify the potential risks (\eg, \emph{a van in the front-left}). (c) Evaluating across PanoVQA, our proposed PLM significantly outperforms all other models across all categories, yielding superior omni-scene understanding. }
    \label{fig:pano_vqa_pipeline}
\end{center}%
}]
{
  \renewcommand{\thefootnote}
    {\fnsymbol{footnote}}
  \footnotetext[2]{This work was done while Weijia Fan was a visiting student at KIT.}
  \footnotetext[1]{Correspondence: \href{mailto:junwei.zheng@kit.edu}{\textcolor{blue}{junwei.zheng@kit.edu}}, \href{mailto:jiamingzhang@hnu.edu.cn}{\textcolor{blue}{jiamingzhang@hnu.edu.cn}}.}
}
\begin{abstract}
Existing vision-language models (VLMs) are tailored for pinhole imagery, stitching multiple narrow field-of-view inputs to piece together a complete omni-scene understanding. Yet, such multi-view perception overlooks the holistic spatial and contextual relationships that a single panorama inherently preserves.
In this work, we introduce the \textbf{Panorama-Language Modeling (PLM)} paradigm, a unified $360^\circ$ vision-language reasoning that is more than the sum of its pinhole counterparts.
Besides, we present \textbf{PanoVQA}, a large-scale panoramic VQA dataset that involves adverse omni-scenes, enabling comprehensive reasoning under object occlusions and driving accidents. To establish a foundation for PLM, we develop a plug-and-play panoramic sparse attention module that allows existing pinhole-based VLMs to process equirectangular panoramas without retraining. Extensive experiments demonstrate that our PLM achieves superior robustness and holistic reasoning under challenging omni-scenes, yielding understanding greater than the sum of its narrow parts.
Project page: \url{https://github.com/InSAI-Lab/PanoVQA}. 
\end{abstract}

\section{Introduction}
\label{sec:intro}
Vision-Language Models (VLMs)~\cite{liu2023llava1.5,li2024llava_onevision,bai2023qwen_vl,chen2024internvl} have recently demonstrated remarkable capabilities in bridging the gap between visual perception and natural language understanding~\cite{liu2023visual_instruction_tuning,li2023blip_2}. Foundation models such as LLaVA~\cite{liu2023visual_instruction_tuning} and BLIP-2~\cite{li2023blip_2} have excelled at a wide range of tasks, including visual question answering (VQA)~\cite{antol2015vqa}, image captioning~\cite{xu2015image_captioning}, and complex visual reasoning~\cite{hudson2019gqa}, paving the way for more general-purpose AI systems.

Despite their success, most existing VLMs are tailored for standard ``pinhole'' imagery, which captures a narrow field-of-view (FoV). This limitation becomes critical when applied to omnidirectional ($360^\circ$) scenes, which are increasingly prevalent in applications like autonomous driving~\cite{yang2021_wildPASS,cao2024oass_blendpass}, robotics~\cite{ying2025mmwalk}, and immersive augmented/virtual reality (AR/VR)~\cite{huang2017VR}. To comprehend a panoramic scene, current methods typically resort to a ``stitching'' approach: sampling multiple narrow-view crops, processing them individually, and attempting to piece together a complete understanding~\cite{wang2025multi}. This multi-view perception, however, fundamentally overlooks the holistic spatial and contextual relationships that a single, unified panorama inherently preserves. It breaks the seamless $360^\circ$ continuity and fails to model the crucial ``wrap-around'' nature of panoramas, where the left and right edges of the image are connected. 

This research gap is exacerbated by two primary challenges. (1) A lack of dedicated, large-scale benchmarks for training and validation. 
Existing datasets are limited to either multi-view pinhole VQA~\cite{qian2024nuscenes_qa,tian2025nuscenes_spatialqa} or panoramic data without corresponding VQA pairs~\cite{yang2021_wildPASS,cao2024oass_blendpass}. Crucially, they lack the diverse, adverse situations needed to test robust real-world reasoning. (2) The inherent incompatibility of model architecture. Equirectangular projections (ERP), the standard $360^\circ$ format, introduce severe geometric distortions and have much higher resolutions than typical pinhole images. Applying narrow-FoV Transformer-based VLMs naively is not only computationally prohibitive, due to the $O(n^2)$ complexity of dense attention, but also fails to model the unique projection topology.

\begin{figure}
    \centering
    \vskip -0.5em
    \includegraphics[width=1.0\linewidth]{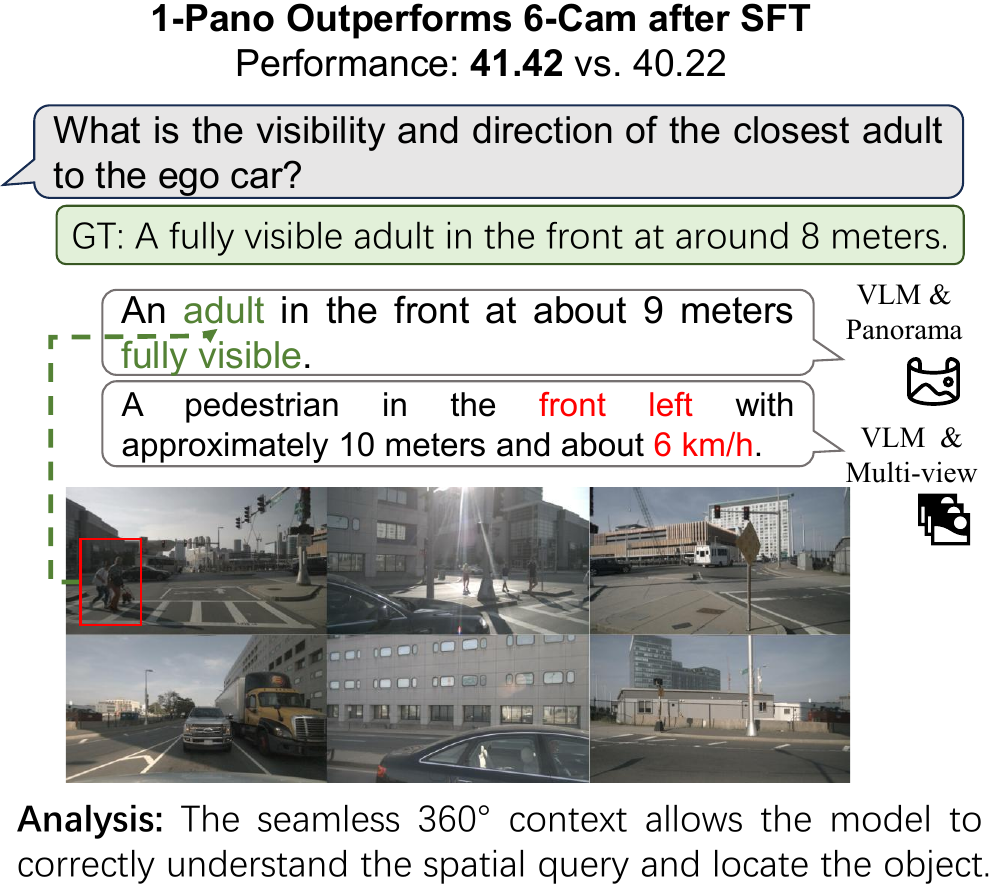}
    \caption{1-Pano (41.42$\%$) outperforms 6-Cam (40.22$\%$) on PanoVQA-mini.
    The panorama's seamless $360^\circ$ context is key for spatial awareness. 
    As shown, the 6-cam model fails the query, e.g., misidentifying the direction.
    In contrast, the 1-Pano model leverages the full context to, e.g., correctly locate the object, matching the GT. More examples can be found in the supplementary.}
    \label{fig:compare_pano_6cam}
    \vspace{-1.5em}
\end{figure}

To address the issues above, this work validates the hypothesis that panorama-language understanding is more than the sum of its pinhole counterparts, as shown in Fig.~\ref{fig:compare_pano_6cam}.
We introduce the \textbf{Panorama-Language Model (PLM)}, a well-designed VLM for panoramas, where we propose a panoramic sparse attention (PSA). This design ensures that our module remains compatible with existing pre-trained VLMs, enabling them to efficiently process $360^\circ$ inputs.
To train and evaluate our PLM and existing VLMs, we present \textbf{PanoVQA}, a new, large-scale panoramic visual question answering dataset. As illustrated in Fig.~\ref{fig:fig1_a}, PanoVQA is distinct in its integration of a vast collection of diverse and adverse omni-scenes. It features complex scenarios involving occlusions, vehicular accidents, and other challenging environmental conditions. This design forces models to move beyond simple object recognition and engage in comprehensive reasoning about the entire complex situation.

To summarize, our contributions are as follows.
\begin{compactitem}
    \item  We introduce PanoVQA, a large-scale dataset curated with diverse ``omni-scenes'', including normal driving, occlusion, and accident scenarios, to facilitate comprehensive $360^\circ$ reasoning.
    \item We propose a panoramic sparse attention mechanism, a novel module specifically designed to address equirectangular distortion and efficiently capture long-range spatial dependencies in panoramic images. 
    \item Extensive experiments demonstrate the limitations of existing VLMs on panoramic tasks and validate the effectiveness and superior reasoning capabilities of our proposed PLM framework.
\end{compactitem}

\begin{table*}[th]
\centering
\caption{
\textbf{Existing multi-view VQA or panoramic datasets}. We compare key attributes: data volume (frames, QAs), input modality (pinhole \textit{vs.} panorama), and scenario diversity. While previous dataset is limited to either pinhole VQA or panoramic data without VQA, our PanoVQA is the first to provide 360$^{\circ}$ images and VQA (653K pairs), spanning \textit{normal}, \textit{occlusion}, and \textit{accidental} scenarios.}
\label{tab:comparsion_pano_vqa}
\setlength{\tabcolsep}{6pt} 
\begin{adjustbox}{width=\textwidth}
\begin{tabular}{@{}l l c c c c c c c@{}}
\toprule
\textbf{Dataset} & \textbf{Year} & \textbf{\#Frames} & \textbf{\#QAs} & \textbf{Input} & \textbf{Real/Syn} & \textbf{Scenario} & \textbf{Occlusion} & \textbf{Accident} \\
\midrule
NuScenes-QA~\cite{qian2024nuscenes_qa} & AAAI'24 & 34K & 460K & Pinhole & Real & Driving & $\times$ & $\times$ \\
DriveLM~\cite{sima2024drivelm} & ECCV'24 & 5K & 443K & Pinhole & Real \& Syn & Driving & $\times$ & $\times$ \\
OmniDrive~\cite{wang2025omnidrive} & CVPR'25 & 34K & 136K & Pinhole & Real \& Syn & Driving & $\times$ & $\times$ \\
NuPlanQA-1M~\cite{park2025nuplanqa} & ICCV'25 & 90K & 1M & Pinhole & Real & Driving & $\times$ & $\times$ \\
NuScenes-SpatialQA~\cite{tian2025nuscenes_spatialqa} & arXiv'25 & 34K & 460K & Pinhole & Real & Driving & $\times$ & $\times$ \\
DeepAccident~\cite{wang2024deepaccident} & AAAI'24 & 57K & $\times$ & Pinhole & Syn & Driving & $\times$ & \checkmark \\
VRU-Accident~\cite{kim2025vru_accident} & ICCVW'25 & - & 6K & Pinhole & Real & Driving & $\times$ & \checkmark \\
ChatBEV~\cite{xu2025chatbev} & arXiv'25 & 25K & 137K & BEV & Real & Driving & $\times$ & $\times$ \\

mmWalk~\cite{ying2025mmwalk} & NeurIPS'25 & 7.5K & 69K & Panorama & Syn & Walking & $\times$ & $\times$ \\
WildPASS~\cite{yang2021_wildPASS} & CVPR'21 & 2.5K & $\times$ & Panorama & Real & Mixed & $\times$ & $\times$ \\
BlendPASS~\cite{cao2024oass_blendpass} & ECCV'24 & 2.1K & $\times$ & Panorama & Real & Driving & \checkmark & $\times$ \\
\rowcolor[gray]{.9} \textbf{PanoVQA (Ours)} & \textbf{-} & \textbf{44.6K} & \textbf{653K} & \textbf{Panorama} & \textbf{Real \& Syn} & \textbf{Driving} & \textbf{\checkmark} & \textbf{\checkmark}\\ 
\bottomrule
\end{tabular}
\end{adjustbox}
\vskip -0.5em
\end{table*}

\section{Related Work}
\label{related_work}

\subsection{Panoramic Scene Benchmark} 
Recent VQA benchmarks target scene understanding, particularly in autonomous driving shown in Table~\ref{tab:comparsion_pano_vqa}. Initial works like NuScenes-QA~\cite{qian2024nuscenes_qa} provide template-based, single-word-answer questions, while NuScenes-SpatialQA~\cite{tian2025nuscenes_spatialqa} focuses on evaluating spatial reasoning. Some benchmarks address more complex cognitive tasks. DriveLM~\cite{sima2024drivelm} introduced Graph VQA to model driver reasoning. Works focusing on planning, like OmniDrive~\cite{wang2025omnidrive}, use counterfactual trajectory reasoning, while NuPlanQA-1M~\cite{park2025nuplanqa} scales this to one million diverse VQAs generated by GPT-4o. A parallel line focuses on pedestrian-centric assistance. mmWalk~\cite{ying2025mmwalk} is designed for pedestrians with blindness or low vision (BLV), capturing data from walker-egocentric, guide-dog, and aerial views. VizWiz-VQA~\cite{gurari2018vizwiz} uses images and spoken questions from blind users. 

However, these benchmarks mainly use forward-facing, multi-camera, or egocentric perspectives. They do not evaluate a model's ability to reason over a complete $360^\circ$ scene from a single panoramic input. Our PanoVQA benchmark is designed to fill this gap, providing a dedicated resource for evaluating holistic outdoor scene understanding.

\subsection{Panoramic Scene Understanding} 
Holistic panoramic scene understanding involves tasks such as 3D layout estimation~\cite{sun2019horizonnet}, pose estimation~\cite{zheng2025spr}, object detection, and depth estimation~\cite{li2025DA2}. Foundational works like PanoContext~\cite{zhang2014panocontext} and DeepPanoContext~\cite{zhang2021deeppanocontext} established the importance of 3D scene context, which is particularly strong in $360^\circ$ images. In the context of panoramic scenes, a primary challenge is the severe geometric distortion in ERP. Models like SphereNet~\cite{coors2018spherenet}, UniFuse~\cite{jiang2021unifuse}, and PanoFormer~\cite{shen2022panoformer} address this by operating on the sphere, fusing different projections, or using distortion-invariant patches. 
Other works~\cite{zhang2022bending_deformable, zhang2024deformable_panorama,zheng2024ops,hu2024deformable} incorporate deformable design to address image distortions and object deformation in panoramas.
A parallel track, dominant in autonomous driving, projects sensor data into a Bird's-Eye-View (BEV). LSS~\cite{philion2020lss} introduced the ``lift-splat'' method for this projection. BEVFormer~\cite{li2022bevformer,yang2023bevformer_v2} advanced this with spatiotemporal transformers, while OneBEV~\cite{wei2024onebev} adapted the paradigm for single panoramas using a Mamba-based state space model. These Transformer-based models are often computationally expensive. In contrast, our work proposes a dynamic sparse attention mechanism where queries learn which tokens to attend to, improving performance for panoramic inputs while maintaining low overhead.



\subsection{Vision-Language Models} 
Recent large Vision-Language Models (VLMs) like LLaVA~\cite{liu2023llava1.5}, Qwen-VL~\cite{bai2023qwen_vl}, InternVL~\cite{chen2024internvl}, and GLM-V~\cite{hong2025glm} have excelled at tasks like Visual Question Answering (VQA)~\cite{antol2015vqa}, captioning~\cite{xu2015image_captioning}, and retrieval~\cite{wu2021fashion} by bridging vision and language. Their application is expanding into specialized domains like autonomous driving and robotics.
However, existing VLMs are designed for standard pinhole images, which offer a fragmented view with discontinuous information. This creates a gap, as they cannot process the rich, seamless context of a $360^\circ$ scene. Our work, PLM, fills this gap by introducing a PSA that operates directly on panoramic inputs, enabling a more complete and grounded understanding of the full $360^\circ$ environment.

\section{Methodology} 
\label{sec:method}

\subsection{PanoVQA Benchmark} 
To address the limitations of existing datasets in evaluating complex, $360^\circ$ driving-scene understanding, we introduce the PanoVQA benchmark. As detailed in our comparison in Table~\ref{tab:comparsion_pano_vqa}, PanoVQA is the first large-scale VQA dataset specifically designed for panoramic inputs that covers a comprehensive range of driving situations. It provides a robust framework for evaluating driving-focused LLMs and Agents by incorporating diverse scenarios, including normal driving, complex occlusions, and high-risk accidents, which are critically underrepresented in previous benchmarks. The overall pipeline for the generation of our PanoVQA dataset is illustrated in Fig.~\ref{fig:fig1_a}.

\noindent \textbf{Categories.} To create PanoVQA for comprehensive evaluation, we designed a total of 12 visual question categories, as detailed in Table~\ref{tab:panovqa_category}. These categories are derived from three distinct datasets to cover a wide range of panoramic visual understanding tasks in autonomous driving. The PanoVQA-N addresses general driving scenarios, tasking models with (N1) Scene Captioning (including object existence, counting, and localization), (N2) Object Identification (focusing on position, distance, and direction), and understanding both (N3) Object-to-Object (O2O) and (N4) Ego-to-Object (E2O) Spatial Relationships. The PanoVQA-O specifically targets complex occlusion scenarios, with questions designed to test reasoning about unseen elements, such as (O1) Occlusion Relationships, (O2) Inferring Actions of Occluded Objects, and (O3) Identifying Actions to Prevent Accidents. Finally, the PanoVQA-D is collision-focused and built for risk analysis, acknowledging that environment, weather, and intersections impact judgment; its tasks include (D1) Environment/Weather assessment, (D2) Collision Risk Assessment, (D3) Severity Assessment, (D4) Action Avoidance planning, and (D5) Time to Collision Estimation.

\begin{figure}
    \centering
    \includegraphics[width=0.7\linewidth]{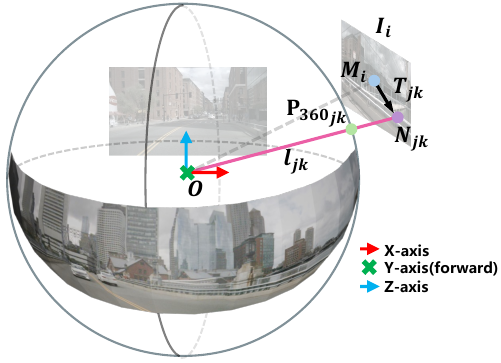}
    \caption{
    Panorama generation overview. Following~\cite{wei2024onebev}, we center a viewing sphere at $O$. For each camera, an image plane $I_i$ is tangent to a concentric sphere at $M_i$ on its optical axis. For each panorama pixel $P_{360_{jk}}$, we cast a ray $l_{jk}$ from $O$, intersect it with the tangent planes, and sample the color at the projected image coordinate $N_{jk}$. Overlaps are resolved by a fixed camera order (``first hit wins''), ensuring consistency without feature matching.}
    \label{fig:stitch}
\end{figure}

\noindent \textbf{Dataset Generation.} We construct our dataset by formatting and cleaning three existing datasets: NuScenes~\cite{caesar2020nuscenes}, BlendPASS~\cite{cao2024oass_blendpass}, and DeepAccident~\cite{wang2024deepaccident}. A critical first step involves generating panoramic images. For NuScenes and DeepAccident, we follow OneBEV~\cite{wei2024onebev} and use a pure-geometry pipeline (Fig.~\ref{fig:stitch}). 
The BlendPASS dataset already provides panoramic images, requiring no stitching. 

Following image preparation, we process the raw annotations from each dataset to create a structured, json-based annotation of each frame. This format allows us to pre-calculate and explicitly store all the object properties required for our targeted VQA tasks. To formally denote each object and its state, we use a colloquial quadruple representation that is both machine-readable and intuitive. An example of this representation is shown in Fig.~\ref{fig:fig1_b}.
The specific preprocessing for each data subset is as follows:

\begin{table}[t]
\centering
\caption{Distribution of 12 QA types on PanoVQA. Tasks are grouped by their three source datasets: NuScenes (normal scenarios), BlendPASS (occlusion reasoning), and DeepAccident (collision accident), along with their respective sample counts.}
\label{tab:panovqa_category}

\setlength{\tabcolsep}{2.4mm}
\resizebox{\columnwidth}{!}{
    \begin{tabular}{c | l l r}
    \toprule
    \textbf{Data.} & \textbf{ID} & \textbf{QA Task Description} & \textbf{\#Samples} \\
    \midrule
    \multirow{4}{*}{\rotatebox[origin=c]{90}{\makecell{Normal\\Scenes}}}
      & N1 & \cellcolor{gray!10}Scene Captioning & 118,101 \\
      & N2 & \cellcolor{gray!10}Object Identification & 144,927 \\
      & N3 & \cellcolor{gray!10}O2O Spatial Relationships & 117,174 \\
      & N4 & \cellcolor{gray!10}E2O Spatial Relationships & 128,793 \\
    \midrule
    \multirow{3}{*}{\rotatebox[origin=c]{90}{\small{\makecell{Occluded\\Scenes}}}}
      & O1 & \cellcolor{gray!10}Occlusion Relationships & 698 \\
      & O2 & \cellcolor{gray!10}Actions of Occluded Objects & 252 \\
      & O3 & \cellcolor{gray!10}Actions to Prevent Accidents & 342 \\
    \midrule
    \multirow{5}{*}{\rotatebox[origin=c]{90}{\makecell{Accident\\Scenes}}}
      & D1 & \cellcolor{gray!10}Environment, Weather & 38,619 \\
      & D2 & \cellcolor{gray!10}Collision Risk Assessment & 28,774 \\
      & D3 & \cellcolor{gray!10}Severity Assessment & 21,270 \\
      & D4 & \cellcolor{gray!10}Action Avoidance  & 29,405 \\
      & D5 & \cellcolor{gray!10}Time2Collision Estimation & 25,559 \\
    \bottomrule
    \end{tabular}
}
\end{table}

\begin{compactitem} 
\item \textbf{PanoVQA-N (NuScenes)} establishes a baseline for panoramic scene understanding under normal driving conditions. The goal is to train and evaluate a VLM's ability to identify objects and understand their spatial relationships to the ego-vehicle. We parse the annotations to calculate each object's position, which we then represent with the quadruple: \texttt{(category, direction, distance, visibility)}. The format was chosen because it provides the fundamental information required for navigation: what an object is, where it is, and how well it can be seen.
\item \textbf{PanoVQA-O (BlendPASS):} We first employ the DA2~\cite{li2025DA2} for pixel-level depth estimation. An object's distance is then determined by averaging the depth values within its polygonal region. The object's position and angle are used to calculate its direction. To determine occlusion relationships, we identify overlapping objects via a polygonal region and compare their lowest $y$-axis pixel coordinates; the object with the smaller $y$-coordinate is designated the ``occluding object'', and the other is the ``occluded object''. This analysis allows us to calculate a ``visibility'' score based on the Intersection over Union (IoU) region. The quadruple is \texttt{(category, direction, distance, visibility)}.
\item \textbf{PanoVQA-D (DeepAccident):} We include environmental conditions, such as weather and road type, which are critical for braking and visibility. As this dataset focuses on collisions and accident scenarios, we calculate objects' relative angle, speed, and distance. Because our input is a static image, it is difficult to infer speed. Therefore, the objects' speed and distance are provided in the question to facilitate understanding. The resulting quadruple is \texttt{(category, direction, distance, speed)}.
\end{compactitem} 
With these pre-calculated and structured scene annotations, we utilize \texttt{GPT-5-mini} with \texttt{reasoning effort} $= \texttt{\{minimal, low\}}$ as a generator to create the question-answer (QA) pairs. The generated dataset then undergoes a final stage of automated machine cleaning and rigorous human evaluation to ensure quality. Further technical details, including the specific formulas used for calculating relative positions and occlusion, as well as the generation prompts, are available in the supplement.

\begin{table}[t]
    \centering
    \caption{\textbf{Comparison of dataset statistics.} PanoVQA introduces over 650K total samples, offering a larger scale and comprehensive question-answer lengths compared to existing benchmarks.}
    \label{tab:avg_word_length_samples_split}
    \setlength{\tabcolsep}{1.8mm}
    \resizebox{\columnwidth}{!}{
        \begin{tabular}{lrrrr}
            \toprule
            \multirow{2}{*}{\textbf{Dataset}} & \multicolumn{2}{c}{\textbf{Samples}} & \multicolumn{2}{c}{\textbf{Average Word Length}} \\
            \cmidrule(lr){2-3} \cmidrule(lr){4-5}
             & \textbf{Train} & \textbf{Val} & \textbf{Question} & \textbf{Answer} \\
            \midrule
            NuScenes~\cite{qian2024nuscenes_qa} & 419,336 & 89,659 & 15.6 & 34.8 \\
            BlendPASS~\cite{cao2024oass_blendpass} & -- & 1,292 & 17.1 & 51.0 \\
            DeepAccident~\cite{wang2024deepaccident} & 119,299 & 24,328 & 24.5 & 41.4 \\ 
            \rowcolor{gray!15} \textbf{PanoVQA} (Ours) & 538,635 & 115,279 & 19.07 & 42.4 \\            
            \bottomrule
        \end{tabular}
    }
    \vskip -1.0em
\end{table}

\noindent \textbf{Filtration and Statistics.} To ensure the quality of the data, we first applied an automatic filtering process using manually defined keywords to remove undesirable QA pairs. Following this refinement, our complete PanoVQA dataset comprises 653K QA pairs, split into $\mathbf{538K}$ for training and $\mathbf{115K}$ for validation based on the original dataset's setting. Furthermore, for efficient evaluation, we curated a smaller version, PanoVQA-mini, containing 25K QA pairs, specifically $\mathbf{17,378}$ for training and $\mathbf{8,184}$ for validation.

\noindent \textbf{Human Evaluation.} To ensure the correctness of the annotation, we conduct mix human and GPT evaluation on a mini version. The result shows our annotations achieve high quality. As shown in Table~\ref{tab:quality_score_opt}, overall scores affirm the dataset's reliability

\begin{table}[t]
\centering
\caption{Data quality verified by the human-GPT evaluation scores. The higher the better, up to 5.  
}
\label{tab:quality_score_opt}
    \setlength{\tabcolsep}{1.19mm}    
    \resizebox{\columnwidth}{!}{
    \begin{tabular}{lcccc}
    \toprule
    \multirow{2}{*}{\textbf{Dataset}} & \textbf{Question} & \multicolumn{3}{c}{\textbf{Answer}} \\
    \cmidrule(lr){3-5}
    & \textbf{Quality} & \textbf{Correctness} & \textbf{Actionability} & \textbf{Fluency} \\
    \midrule
    PanoVQA-N & 4.73 & 4.50 & 4.31 & 4.84 \\
    PanoVQA-O & 4.21 & 4.04 & 3.52 & 4.66 \\
    PanoVQA-D & 4.80 & 4.72 & 4.64 & 4.91 \\
    \bottomrule
    \end{tabular}
    }
    \vskip -1.5em
\end{table}

\subsection{Preliminary of narrow-FoV VLMs}
The original Vision Transformer (ViT) processes an image by first dividing it into a sequence of non-overlapping, fixed-size patches. These patches are flattened and then linearly projected into vector embeddings. A learnable $\texttt{cls}$ token embedding is prepended to this sequence, and positional embeddings are added to all tokens to retain spatial information. This results in the final input sequence $\mathbf{h}^{(0)} \in \mathbb{R}^{L \times d}$, where $L$ is the number of patch tokens (include  $\texttt{cls}$ token) and $d$ is the hidden dimension.

Each Transformer block contains a Multi-Head Self-Attention (MSA) and a Feed-Forward Network (FFN). The computation for the $l$-th block, including residual connections and Layer Normalization (LN), which is expressed as
\vskip -1.5em 
\begin{align} 
    \tilde{\mathbf{h}}^{(l)} &= \text{LN}\left(\text{MSA}^{(l)}(\mathbf{h}^{(l-1)}) + \mathbf{h}^{(l-1)}\right), \label{eq:msa} \\ \mathbf{h}^{(l)} &= \text{LN}\left(\text{FFN}^{(l)}(\tilde{\mathbf{h}}^{(l)}) + \tilde{\mathbf{h}}^{(l)}\right). 
\end{align}
\vskip -0.5em

The MSA in Eq.~\ref{eq:msa} is a variant of self-attention that runs multiple independent attention heads in parallel. It computes attention through a weighted sum of Value ($\mathbf{V}$) vectors based on attention scores derived from Query ($\mathbf{Q}$) and Key ($\mathbf{K}$) vectors. For the $l$-th layer, the projected matrices of the input hidden state $\mathbf{h}^{(l-1)}$ are defined as $\mathbf{Q}^{(l)}=\mathbf{h}^{(l-1)}\mathbf{W}_Q^{(l)}$,  $\mathbf{K}^{(l)}=\mathbf{h}^{(l-1)}\mathbf{W}_K^{(l)}$, and  $\mathbf{V}^{(l)}=\mathbf{h}^{(l-1)}\mathbf{W}_V^{(l)}$.
The computation for a self-attention head is calculated as:
\vskip -0.5em
\begin{align}
    \text{Attention}\left( \mathbf{Q},\mathbf{K},\mathbf{V}\right) = \text{softmax}\left( \mathbf{Q}\mathbf{K}^T/\sqrt{d} \right) \mathbf{V}.
\end{align}

The $\text{MSA}^{(l)}$ is formed by concatenating the results from multiple heads and applying an output projection $\mathbf{W}_O^{(l)}$.

In the standard dense self-attention, the query of every token is compared against the key of all other tokens. This mechanism, while effective, introduces a significant bottleneck, \ie, its computational and memory complexity scales quadratically $O(L^2)$ with the number of tokens $L$, making it prohibitively expensive for panoramic inputs.
While Simplified Sparse Attention (SSA) has proposed to reduce cost, it risks losing important information by pre-determining an inflexible attention structure.
Thus, we propose a hybrid attention mechanism that combines local and global sparse attention. This approach is designed to efficiently capture both fine-grained local patterns and critical long-range spatial dependencies within the panoramic image, balancing computational efficiency with model performance.

\begin{figure*}[t]
    \captionsetup{type=figure}
    \centering
    \begin{subfigure}[t]{0.32\textwidth}
        \centering
        \includegraphics[width=\textwidth]{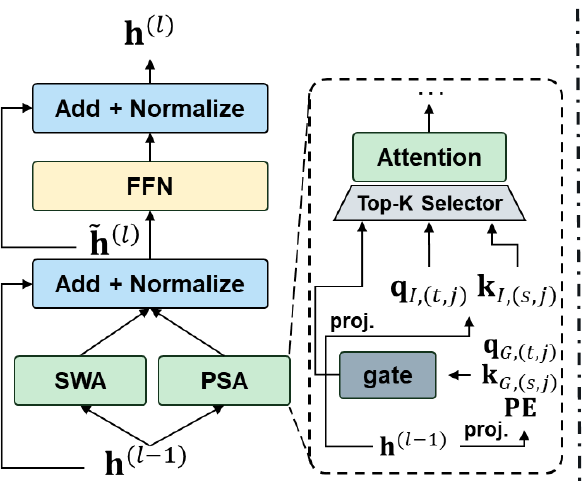}
        \caption{Our panorama sparse attention block.}
        \label{fig:fig4_a}
    \end{subfigure}
    \begin{subfigure}[t]{0.68\textwidth}
        \centering
        \includegraphics[width=0.98\textwidth]{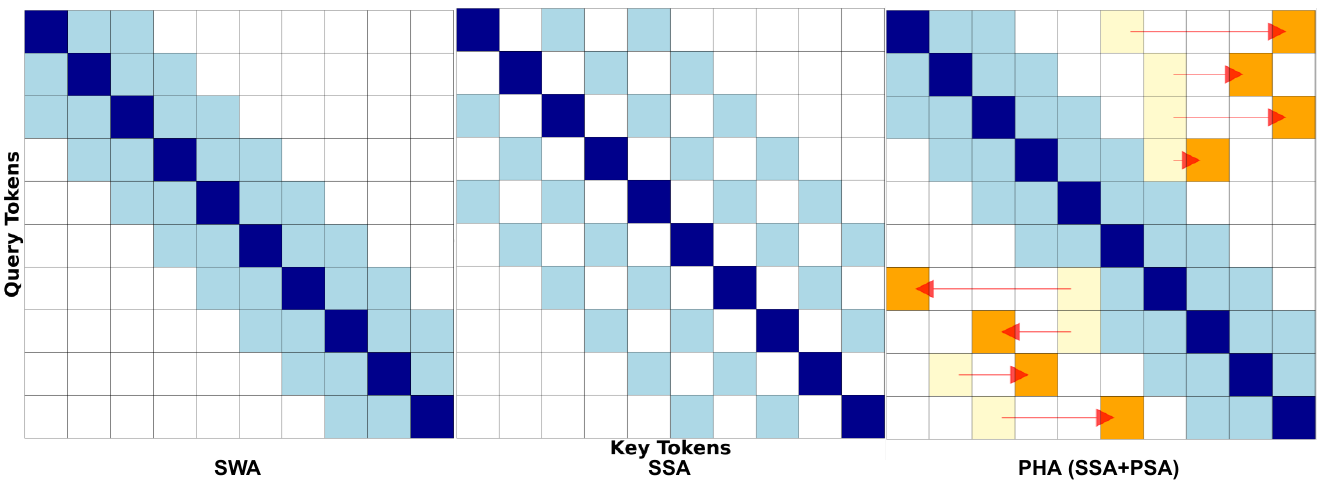}
        \caption{Attention mask visualization.}
        \label{fig:fig4_b}
    \end{subfigure}
  \caption{\textbf{Left}: Structure of our proposed attention block with SWA and PSA. \textbf{Right}: The visualization of attention masks for Sliding Window Attention (SWA), Simplified Sparse Attention (SSA), and Panoramic Sparse Attention (PSA), respectively.}
  \label{fig:attention_mask_vis}
  \vskip -1.0em
\end{figure*}

\subsection{Panorama-Language Model}
The overview of our panorama-language model (PLM) is presented in Fig.~\ref{fig:attention_mask_vis}. 
Similar to modern VLMs~\cite{liu2023llava1.5, li2024llava_onevision, bai2023qwen_vl}, our PLM is composed of three components: (1) a panorama-enhanced ViT, (2) an MLP-based merger, and (3) a large language model (LLM). 
To capture the fine-grained local patterns and long-range spatial relationships in a panoramic image, we modify the standard ViT by adding local and global attention mechanisms in a parallel manner, which preserves the existing structure of VLMs.

\noindent \textbf{Sliding Window Attention.} The local head is implemented using Sliding Window Attention (SWA)~\cite{child2019window_attention}, where we set the sliding stride equal to the window size, resulting in $W$ non-overlapping windows. The input sequence $\mathbf{h}^{(l-1)}$ is partitioned into $W$ non-overlapping windows, $\{\mathbf{h}_1, \dots, \mathbf{h}_W\}$, where each window $\mathbf{h}_w \in \mathbb{R}^{L_w \times d}$ and $L_w$ is the window size ($L_w \ll L$). Self-attention is then computed independently within each window $w \in \{1, \dots, W\}$. The Query, Key, and Value matrices are projected from the windowed input $\mathbf{h}_w$, \eg, 
$\mathbf{Q}_{w}^{(l)} = \mathbf{h}_{w}\mathbf{W}_{Q,\text{local}}^{(l)}, \quad \mathbf{K}_{w}^{(l)} = \mathbf{h}_{w}\mathbf{W}_{K,\text{local}}^{(l)},\quad \mathbf{V}_{w}^{(l)} = \mathbf{h}_{w}\mathbf{W}_{V,\text{local}}^{(l)},$
where $\mathbf{Q}_{w}^{(l)}, \mathbf{K}_{w}^{(l)}, \mathbf{V}_{w}^{(l)} \in \mathbb{R}^{L_w \times d}$ are the query, key, and value matrices for window $w$, and $L_w$ is the window size ($L_w \ll L$). A window attention is calculated by
\begin{align}
    \mathbf{o}_{w}^{(l)} &= \text{Attention} \left( \mathbf{Q}_{w}^{(l)}, \mathbf{K}_{w}^{(l)},\mathbf{V}_{w}^{(l)}  \right).
\end{align}

The output for the SWA is obtained by reassembling the outputs from all windows and applying a projection $\mathbf{W}_{O,\text{local}}^{(l)}$:
\begin{align} 
    \mathbf{Attn}_{\text{local}}^{(l)} &= \text{Reassemble}(\mathbf{o}_{1}^{(l)}, \dots, \mathbf{o}_{W}^{(l)}), \\
    \text{SWA}(\mathbf{h}^{(l-1)}) &= \mathbf{Attn}_{\text{local}}^{(l)} \cdot \mathbf{W}_{O,\text{local}}^{(l)}. \label{eq:local_attn} 
\end{align}

This local attention mechanism reduces the complexity from $O(L^2)$ to $O(W \cdot L_w^2) = O(L \cdot L_w)$, but it prevents tokens from different windows from interacting.

\noindent \textbf{Panoramic Sparse Attention.} 
The global head is implemented by Panoramic Sparse Attention (PSA), which dynamically selects the Top-K most relevant key tokens for each query token. 
First, a dynamic selector module computes a non-causal score matrix $\mathbf{I} \in \mathbb{R}^{L \times L}$. The score $I_{t,s}$, representing the relevance between a query token $t$ and a key token $s$, is calculated as:
\vskip -0.5em
\begin{align}
  I_{t,s} = \sum_{j=1}^{H^I} \Big( 
      &\operatorname{gate}(\mathbf{q}_{G,(t,j)}, \mathbf{k}_{G,(s,j)}, t, s) \nonumber \\
      &\cdot \operatorname{ReLU}(\mathbf{q}_{I,(t,j)}^\top \mathbf{k}_{I,(s,j)}) 
  \Big), \label{eq:score} \\
\
  \operatorname{gate}(\mathbf{q}, \mathbf{k}, t, s) = 
      &\operatorname{sigmoid}\left(
          \operatorname{MLP}\left( \mathbf{q}^\top \mathbf{k} + \text{PE}_{t,s} \right) 
      \right), \label{eq:gate}
\end{align}
where $H^I$ is the number of selector heads. For each head $j$, the vectors $\mathbf{q}_{G,(t,j)}, \mathbf{k}_{G,(s,j)} \in \mathbb{R}^{d_G}$ and $\mathbf{q}_{I,(t,j)}, \mathbf{k}_{I,(s,j)} \in \mathbb{R}^{d_I}$ are derived from the input tokens $\mathbf{h}_t^{(l-1)}$ and $\mathbf{h}_s^{(l-1)}$ through separate linear projections ($d_G \ll d, d_I \ll d$). The $\text{gate}(\cdot)$ network in Eq.~(\ref{eq:gate}) is an MLP that incorporates a learnable position embedding ($\text{PE}_{t,s}$) to make the selection process position-aware.

Consequently, we compute the main sparse attention projections from the input $\mathbf{h}^{(l-1)}$, \eg, 
$
\mathbf{Q}_{S}^{(l)} = \mathbf{h}^{(l-1)}\mathbf{W}_{Q,\text{sparse}}^{(l)}, \quad
\mathbf{K}_{S}^{(l)} = \mathbf{h}^{(l-1)}\mathbf{W}_{K,\text{sparse}}^{(l)}, \quad
\mathbf{V}_{S}^{(l)} = \mathbf{h}^{(l-1)}\mathbf{W}_{V,\text{sparse}}^{(l)}.
$ 
 For each query token $t$ (row $\mathbf{q}_t^{(l)}$ from $\mathbf{Q}_{S}^{(l)}$), we identify the set of Top-K key indices $\mathcal{S}_t = \{s \mid I_{t,s} \in \text{Top-K}(\mathbf{I}_{t,:}) \}$. The sparse attention output is computed only over these selected key-value pairs:
\begin{align}
\mathbf{o}_t^{(l)} = \text{Attention}\left( \mathbf{q}^{(l)}_t, \mathbf{K}_{\mathcal{S}_t}^{(l)}, \mathbf{V}_{\mathcal{S}_t}^{(l)} \right),
\end{align}
where $\mathbf{K}_{\mathcal{S}_t}^{(l)}$ and $\mathbf{V}_{\mathcal{S}_t}^{(l)}$ are matrices formed by stacking the key and value vectors from $\mathbf{K}_{S}^{(l)}$ and $\mathbf{V}_{S}^{(l)}$ corresponding to the indices in $\mathcal{S}_t$.
The final output for the PSA module is obtained by stacking all token outputs $\mathbf{o}_t^{(l)}$ and applying an output projection $\mathbf{W}_{O,\text{sparse}}^{(l)}$:
\begin{align}
\mathbf{Attn}_{\text{sparse}}^{(l)} &= \text{Stack}(\mathbf{o}_1^{(l)}, \dots, \mathbf{o}_{L}^{(l)}), \\
\text{PSA}(\mathbf{h}^{(l-1)}) &= \mathbf{Attn}_{\text{sparse}}^{(l)} \mathbf{W}_{O,\text{sparse}}^{(l)}. \label{eq:psa_output}
\end{align}
\vskip -0.5em
We define Panoramic Hybrid Attention (PHA) as the parallel combination of SWA and PSA. The $l$-th block calculation is reformulated as:
\begin{small}
    \begin{align}
    \tilde{\mathbf{h}}^{(l)} &= \text{LN}\left( \text{SWA}(\mathbf{h}^{(l-1)}) + \text{PSA}(\mathbf{h}^{(l-1)}) + \mathbf{h}^{(l-1)} \right), \label{eq:hybrid_attn} \\
    \mathbf{h}^{(l)} &= \text{LN}\left(\text{FFN}^{(l)}(\tilde{\mathbf{h}}^{(l)}) + \tilde{\mathbf{h}}^{(l)}\right). \label{eq:hybrid_ffn}
    \end{align}
\end{small}

Fig.~\ref{fig:fig4_b} visualizes three attention masks. Sliding Window Attention (SWA) (left) captures fine-grained local patterns (blue diagonal band) but fails to model the global context. Simplified Sparse Attention (SSA) (middle) models long-range dependencies, but its fixed pattern is not optimized for the unique structure of panoramas. 
Our proposed Panoramic Hybrid Attention (PHA) (right) is specifically designed to overcome these limitations. 
PHA retains SWA's local window to capture fine-grained details while dynamically incorporating long-range dependencies (visualized as yellow-to-orange points) to connect distant token, such as front-back sides, or the left-right sides. Additionally, PSA utilizes patch-wise deformable attention and gating mechanisms to effectively handle the object distortion and filter out uninformative areas inherent in panoramic images. Further discussion is provided in the supplement.




\begin{table*}[t]
\centering
\caption{Results for all models on the PanoVQA benchmark. `-' means not available. The metric is the normalized GPT-score. All experiments are conducted on the PanoVQA.}
\vskip -0.5em
\label{tab:comparison_of_exisiting_and_PLM}
\resizebox{\textwidth}{!}{    
\setlength{\tabcolsep}{1.3mm}{
    \begin{tabular}{lcccccccccccccc}
    \toprule
    \multirow{2}{*}{\textbf{Model}} & \multicolumn{4}{c}{\textbf{PanoVQA-N}} & \multicolumn{3}{c}{\textbf{PanoVQA-O}} & \multicolumn{5}{c}{\textbf{PanoVQA-D}} & \multirow{2}{*}{\textbf{Avg.}} \\
    \cmidrule{2-13}
    & \textbf{N1} & \textbf{N2} & \textbf{N3} & \textbf{N4} & \textbf{O1} & \textbf{O2} & \textbf{O3} & \textbf{D1} & \textbf{D2} & \textbf{D3} & \textbf{D4} & \textbf{D5} & \\
    \midrule
    \rowcolor{gray!15}\multicolumn{14}{l}{\textit{Zero-shot}} \\
    \quad Chameleon-7B~\cite{zhang2025chameleon} & 9.87 & 18.04 & 12.60 & 10.62 & 6.98 & 33.04 & 38.96 & 8.71 & 23.65 & 35.01 & 30.76 & 11.69 & 19.99 \\
    \quad LLaVA-NEXT-7B~\cite{liu2024llavanext} & 11.19 & 17.15 & 15.06 & 12.13 & 10.54 & 40.47 & 46.91 & 8.93 & 20.66 & 36.73 & 23.33 & 18.57 & 21.81 \\
    \quad LLaVA-OneVision-7B~\cite{li2024llava_onevision} & 12.54 & 19.61 & 19.83 & 14.49 & 11.97 & 40.66 & 43.25 & 12.05 & 25.88 & 47.09 & 34.75 & 43.00 & 27.09 \\
    
    \quad Qwen2.5-VL-7B-Ins.~\cite{Qwen2_5_vl2025} & 13.65 & 18.53 & 22.08 & 14.72 & 15.13 & 35.89 & 44.68 & 12.86 & 30.97 & 48.86 & 36.93 & 62.90 & 29.77 \\
    \quad InternVL2-8B~\cite{team2024internvl2} & 12.79 & 22.55 & 24.17 & 15.59 & 9.17 & 43.35 & 50.80 & 15.11 & 33.88 & 49.73 & 46.71 & 40.51 & 30.36 \\
    \quad InternVL3-8B-Ins.~\cite{zhu2025internvl3} & 15.06 & 22.79 & 23.99 & 16.14 & 9.32 & 45.91 & 53.38 & 15.06 & 36.06 & 57.87 & 52.98 & 65.17 & 34.48 \\
    \quad GLM-4-9B~\cite{glm2024chatglm} & 11.73 & 15.6 & 19.36 & 13.34 & 11.93 & 44.74 & 53.44 & 10.93 & 31.42 & 46.64 & 43.29 & 52.36 & 29.57 \\
    \quad DeepSeek-VL2-Small~\cite{lu2024deepseek_vl} & 12.50 & 19.45 & 15.66 & 13.71 & 12.46 & 38.69 & 40.64 & 13.77 & 33.48 & 52.01 & 45.83 & 10.04 & 25.69 \\
    \quad Qwen2.5-VL-32B-Ins.~\cite{Qwen2_5_vl2025} & 16.51 & 20.88 & 26.34 & 17.51 & 9.85 & 46.23 & 54.61 & 15.45 & 35.56 & 54.11 & 55.14 & \textbf{74.55} & 35.56 \\ \midrule

    \rowcolor{gray!15}\multicolumn{14}{l}{\textit{Proprietary}} \\
    \quad Grok-3-Mini & 3.38 & 6.88 & 11.61 & 8.42 & 5.37 & 34.23 & 34.58 & 7.91 & 19.30 & 35.00 & 26.15 & 68.05 & 21.74 \\
    \quad Gemini-2.5-Flash-Lite & 18.83 & 23.61 & 25.21 & 19.07 & 11.21 & 36.81 & 45.25 & 14.08 & 31.72 & 32.62 & 41.79 & 68.64 & 30.74 \\ \midrule
    
    \rowcolor{gray!15}\multicolumn{14}{l}{\textit{Panoramic fine-tuned}} \\
    \quad Qwen2.5-VL-7B & \underline{33.52} & \textbf{34.22} & \underline{27.57} & \underline{27.66} & \textbf{22.66} & \underline{50.00} & \underline{58.26} & \underline{23.21} & \underline{42.08} & \underline{74.07} & \underline{74.90} & \underline{74.34} & \underline{45.21} \\
    \quad PanoLM-7B (Ours) & \textbf{34.79} & \underline{33.54} & \textbf{28.18} & \textbf{28.54} & \underline{18.62} & \textbf{50.21} & \textbf{60.09} & \textbf{23.65} & \textbf{42.92} & \textbf{75.72} & \textbf{76.74} & 72.75 & \textbf{45.48} \\
    \hline \bottomrule
    \end{tabular}
}}
\vskip -0.1em
\end{table*}

\section{Experiments}

\subsection{Implementation Details} 
\noindent \textbf{Model.} Our experiments utilize the Qwen2.5-VL series as the base architecture. Since Qwen2.5-VL already incorporates SWA, our primary modification is the integration of our proposed PSA module. We initialize all model weights from the publicly available checkpoints and conduct full-parameter supervised fine-tuning (SFT), unfreezing and training all parameters on our dataset.

\noindent \textbf{Training \& Evaluation Strategy.} We employ a tiered data strategy contingent on model scale. For large-scale models (7B and larger), we create a balanced SFT training mixture by sampling 90\% of the DeepAccident data and 60\% of the NuScenes data; these models are then evaluated on the complete PanoVQA test set. To mitigate computational costs during the analysis and iterative development of our proposed model, we conduct both training and evaluation on the PanoVQA-mini subset. This methodology is supported by established research on scaling laws~\cite{kaplan2020scaling_law,hoffmann2022training_scaling_law}, which demonstrates that model performance scales predictably with dataset size. Our specific scaling law analyses are presented in Sec.~\ref{sec:scaling_law}.

\noindent \textbf{Metrics.} The curated dataset involves normal, occlusion, and accident scenarios. In our experiments, following LLM as-a-judge paradigm~\cite{li2024llms,li2025preference}, we employ \texttt{gpt-4o-mini} to evaluate the similarity and correctness of the generated answers and the ground truth answers. Following a unified scoring prompt in previous work~\cite{ying2025mmwalk} for the evaluation of all models. For each inference result, the metric is a scaled score $s_i$ (1-5, with 5 being the highest score). To visualize the differences in model performance, we normalized all scores to 1$\sim$100 by $\text{S}_{\text{GPT}}=1/N \times \sum_{i=1}^{N} (s_i -1) / 4 * 100\%$, where $N$ denotes the number of all samples. 


\noindent \textbf{Comparison of VLMs.} To establish a comprehensive baseline, we evaluate a wide range of recent Vision-Language Models (VLMs) on our PanoVQA benchmark. This includes leading open-source models such as Chameleon~\cite{zhang2025chameleon}, LLaVA~\cite{liu2023llava1.5, li2024llava_onevision}, Qwen-VL~\cite{bai2023qwen_vl}, InternVL~\cite{chen2024internvl}, GLM~\cite{hong2025glm}, and DeepSeek-VL~\cite{lu2024deepseek_vl}, as well as high-performing proprietary commercial models, including Grok and the Gemini\cite{team2023gemini}. The results are shown in Table~\ref{tab:comparison_of_exisiting_and_PLM}.

\subsection{Results of PanoVQA}
The comprehensive benchmark results on the PanoVQA test set are presented in Table~\ref{tab:comparison_of_exisiting_and_PLM}. Our evaluation provides a broad comparison between existing models in a zero-shot setting and our models fine-tuned on panoramic data.

\noindent \textbf{Zero-Shot Baseline Performance.} 
Among open-source VLMs, performance generally scales with model size and architectural refinement. The strongest model is Qwen2.5-VL-32B-Ins., which achieves the highest average score of $35.56\% $. In the 7B-9B parameter class, which serves as a direct comparison for our model, InternVL3-8B-Ins sets the top mark at $34.48\% $. This significantly outperforms other models in its class, such as InternVL2-8B ($30.36\% $) and the original Qwen2.5-VL-7B-Ins. ($29.77\% $).

\noindent \textbf{Proprietary Model Performance.} The proprietary models show highly varied results. Gemini-2.5-Flash-Lite delivers a competitive score of 30.74\%. Conversely, Grok-3-Mini struggles with the panoramic task, registering a low score of 21.74\%, which falls behind most 7B-scale open-source competitors. Additional proprietary model evaluations on PanoVQA-mini are provided in the supplementary. Ultimately, the general sub-optimal performance across both open-source and proprietary models highlights a shared struggle with panoramic tasks, underscoring the necessity of our PLM.

\noindent \textbf{Panoramic Fine-Tuned Models.} We compare our PanoLM-7B (PLM) and base model, Qwen2.5-VL-7B, after panoramic SFT. Qwen2.5-VL-7B achieves an average score of $45.21\%$, while our PLM achieves a comparable $45.91\%$, surpassing all other methods.

\subsection{Analysis of PLM}

\noindent \textbf{Ablation study.} 
In Table~\ref{tab:ablation_study}, we conduct an ablation study on the PanoVQA-mini dataset using the Qwen-2.5-3B model to verify the effectiveness of our proposed PSA. For these experiments, we first freeze the LLM and train only the vision components. 
Our baseline (line 1) using LoRA~\cite{hu2022lora} achieves $28.55\%$ on average with 259.21M trainable parameters. As shown in line 3, the model with SSA achieves an average score of $29.80\%$, surpassing the baseline while using significantly fewer parameters (85.09M). This result confirms the effectiveness of using sparse attention for panorama inputs. Next, we evaluate our proposed PSA, which enhances SSA by using a dynamic gate network. With the LLM frozen (line 4), the model with our PSA achieves $32.14\%$, which significantly outperforms both the LoRA baseline ($28.55\%$) and a standard SFT approach with a frozen LLM ($29.34\%$, line 7), despite using far fewer trainable parameters (95.56M vs. 668.68M). Finally, when we unfreeze the LLM (line 5), our PSA-enabled model's performance jumps to $41.49$. This result is comparable to, and even slightly surpasses, the performance of a fully fine-tuned 3B model ($41.42$, line 8). These results demonstrate the effectiveness and parameter efficiency of our proposed PSA method. The SSA can be found in the supplement.

\begin{table}[t]
    \centering
    \caption{Ablation study of components in PanoLM. SSA denotes Simplified Sparse Attention, where we use an MLP to replace our gate network. Experiments are conducted in PanoVQA-mini.} 
    \label{tab:ablation_study}
    \setlength{\tabcolsep}{1.6mm}
    \resizebox{\columnwidth}{!}{
        \begin{tabular}{l cccc r r}
        \toprule
        \multirow{2}{*}{\textbf{Model}} & \multicolumn{4}{c}{\textbf{Training}} & \multirow{2}{*}{\textbf{\makecell{\#Param}}} & \multirow{2}{*}{\textbf{Avg}} \\
        \cmidrule(lr){2-5} 
         & \textbf{ViT} & \textbf{MLP} & \textbf{Module} & \textbf{LLM} & & \\
        \midrule
        3B-Lora & \cmark & \cmark & - & $\times$ & 259.21M & 28.55 \\ \midrule
        
        \multicolumn{7}{l}{\textit{Ablation experiments}} \\
        \quad + SSA & $\times$ & \cmark & \cmark & $\times$ & 85.09M & 29.80 \\
        \quad + PSA & $\times$ & \cmark & \cmark & $\times$ & 95.56M & 32.14 \\
        \quad + PSA & $\times$ & \cmark & \cmark & \cmark & 3196.25M & \textbf{41.49}  \\\midrule
        
        \multicolumn{7}{l}{\textit{SFT}} \\
        \quad 3B & \cmark & \cmark & - & $\times$ & 668.68M & 29.34 \\
        \quad 3B & \cmark & \cmark & - & \cmark & 3754.62M & 41.42 \\
        \bottomrule
        \end{tabular}
}
\end{table}

\begin{figure}[t]
  \centering
  \includegraphics[width=1.0\linewidth]{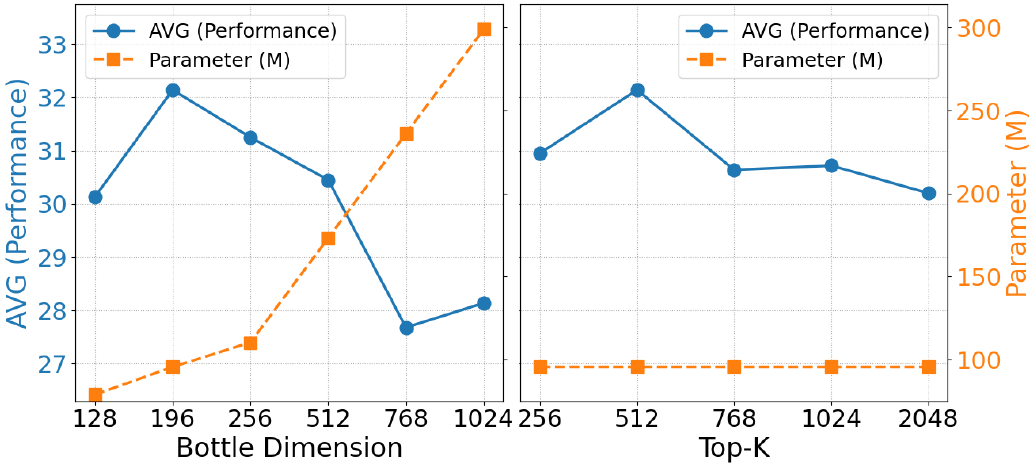}
  \caption{Analysis of the performance-parameter trade-off. \textbf{Left}: Impact of varying bottle dimensions. \textbf{Right}: Impact of varying the selected Top-K.}
  \label{fig:parameter_study}
\end{figure}

\begin{figure}[t]
  \centering
  \includegraphics[width=0.78\linewidth]{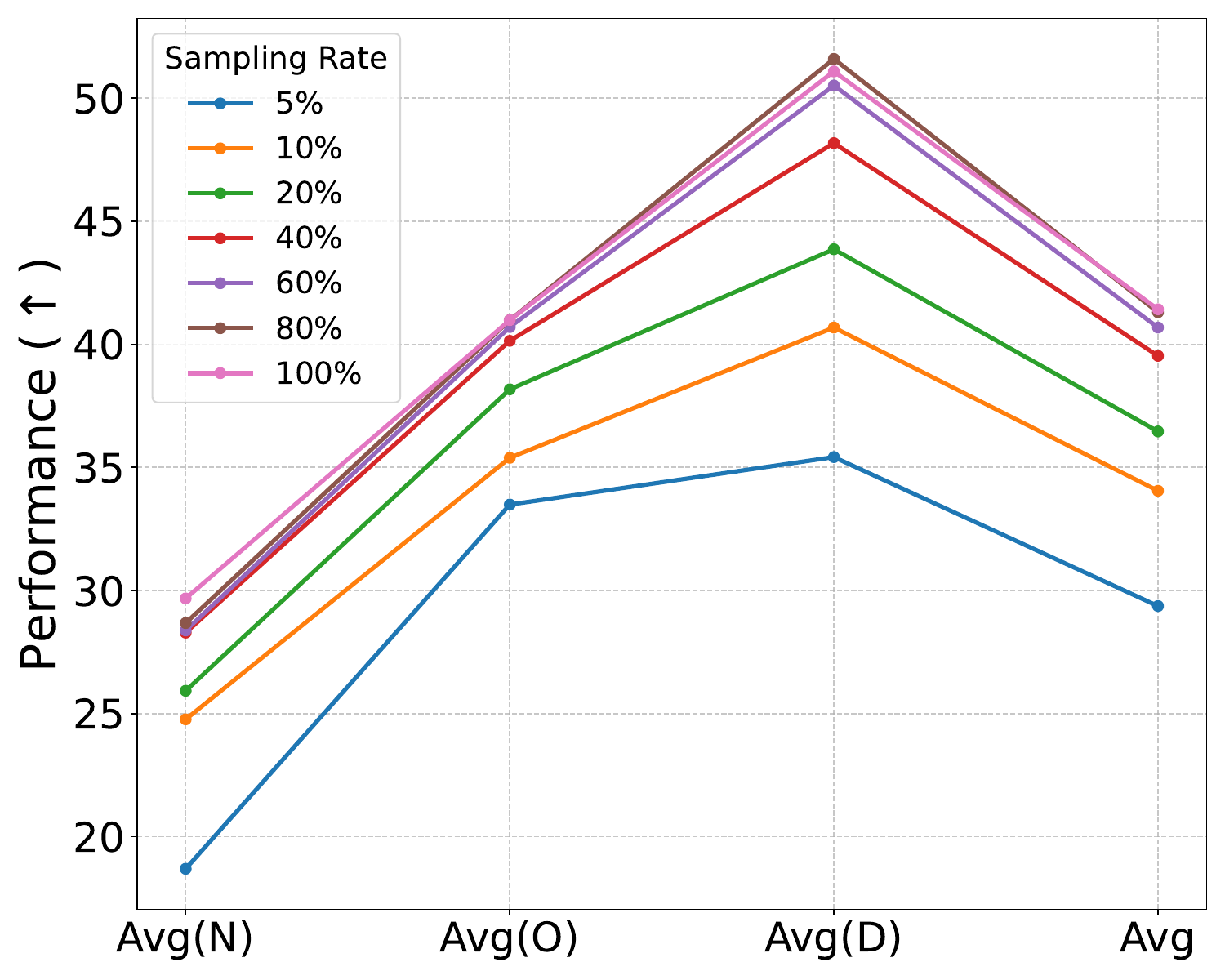}
  \vskip -0.5em
  \caption{Scaling law study on PanoVQA-mini. }
  \vskip -1.0em
  \label{fig:scaling_law}
\end{figure}

\noindent \textbf{Parameters study.}\label{sec:parameters_study}
We analyzed the impact of the bottleneck dimension and Top-K value, plotting performance (blue line) against the number of trainable parameters (orange line). As shown in the left figure, while model parameters (orange) increase with the bottleneck dimension, performance (blue) peaks at 32.10\% with a dimension of 196 before declining. The right figure shows that the Top-K value does not affect parameter count (flat orange line), and performance reaches its peak when $K$=512. Therefore, we adopt a bottleneck dimension of 196 and a Top-K of 512 for all subsequent experiments. All experiments are conducted using PanoLM-3B on PanoVQA-mini.

\noindent \textbf{Scaling law study.}\label{sec:scaling_law}
We confirm the scaling law through experiments on PanoVQA-mini. The results in Fig.~\ref{fig:scaling_law} show that performance scales smoothly with data volume, and the relative gains diminish as data size increases (\eg, from an average score of $40.68\%$ at 60\% data to $41.42\%$ at 100\%). This validates that a smaller, representative subset like PanoVQA-mini is sufficient for reliable model comparison and rapid experimental iteration.

\section{Conclusion}
In this work, we introduce Panorama-Language Modeling (PLM) for holistic panoramic understanding. We propose Panorama Sparse Attention (PSA) to efficiently process $360^\circ$ imagery while maintaining compatibility with pre-trained VLMs. To support this research, we create PanoVQA, a large-scale benchmark for panoramic visual question answering featuring diverse real-world scenarios, including normal, occluded and accidental cases. Extensive experiments demonstrate that our PLM approach effectively addresses the limitations of narrow-FoV multi-view stitching methods and establishes new state-of-the-art performance in panoramic visual reasoning.

\clearpage

\section*{Acknowledgements}
This work was supported by the Shenzhen University Overseas Exchange Scholarship, which supported Weijia Fan's living in Karlsruhe, Germany. This work was supported in part by National Natural Science Foundation of China under Grant No. 62503166, 62576217 and 62576216; Guangdong Provincial Key Laboratory under Grant 2023B1212060076, and also supported by the Intelligent Computing Center of Shenzhen University; and in part by Helmholtz Association of German Research Centers, in part by the Ministry of Science, Research and the Arts of Baden-W\"urttemberg (MWK) through the Cooperative Graduate School Accessibility through AI-based Assistive Technology (KATE) under Grant BW6-03, and in part by the Helmholtz Association Initiative and Networking Fund on the HAICORE@KIT and HOREKA@KIT partition. 

{\small
\bibliographystyle{ieee_fullname}
\bibliography{egbib}

@STRING{aaai	= {AAAI} }

@STRING{cvpr	= {CVPR} }

@STRING{eccv	= {ECCV} }

@STRING{iccv	= {ICCV} }

@STRING{iclr	= {ICLR} }

@inproceedings{caesar2020nuscenes,
  title={nuscenes: A multimodal dataset for autonomous driving},
  author={Caesar, Holger and Bankiti, Varun and Lang, Alex H and Vora, Sourabh and Liong, Venice Erin and Xu, Qiang and Krishnan, Anush and Pan, Yu and Baldan, Giancarlo and Beijbom, Oscar},
  booktitle={Proceedings of the IEEE/CVF conference on computer vision and pattern recognition},
  pages={11621--11631},
  year={2020}
}

@misc{liu2024llavanext,
  title={Llavanext: Improved reasoning, ocr, and world knowledge},
  author={Liu, Haotian and Li, Chunyuan and Li, Yuheng and Li, Bo and Zhang, Yuanhan and Shen, Sheng and Lee, Yong Jae},
  year={2024}
}

@article{zhang2025chameleon,
  title={Chameleon: Fast-slow Neuro-symbolic Lane Topology Extraction},
  author={Zhang, Zongzheng and Li, Xinrun and Zou, Sizhe and Chi, Guoxuan and Li, Siqi and Qiu, Xuchong and Wang, Guoliang and Zheng, Guantian and Wang, Leichen and Zhao, Hang and others},
  journal={arXiv preprint arXiv:2503.07485},
  year={2025}
}

@inproceedings{wu2021fashion,
  title={Fashion iq: A new dataset towards retrieving images by natural language feedback},
  author={Wu, Hui and Gao, Yupeng and Guo, Xiaoxiao and Al-Halah, Ziad and Rennie, Steven and Grauman, Kristen and Feris, Rogerio},
  booktitle={Proceedings of the IEEE/CVF Conference on computer vision and pattern recognition},
  pages={11307--11317},
  year={2021}
}

@article{li2022bevformer,
  title={BEVFormer: Learning Bird’s-Eye-View Representation from Multi-Camera Images via Spatiotemporal Transformers},
  author={Li, Zhiqi and Wang, Wenhai and Li, Hongyang and Xie, Enze and Sima, Chonghao and Lu, Tong and Qiao, Yu and Dai, Jifeng},
  journal={arXiv preprint arXiv:2203.17270},
  year={2022}
}

@inproceedings{yang2023bevformer_v2,
  title={Bevformer v2: Adapting modern image backbones to bird's-eye-view recognition via perspective supervision},
  author={Yang, Chenyu and Chen, Yuntao and Tian, Hao and Tao, Chenxin and Zhu, Xizhou and Zhang, Zhaoxiang and Huang, Gao and Li, Hongyang and Qiao, Yu and Lu, Lewei and others},
  booktitle={Proceedings of the IEEE/CVF conference on computer vision and pattern recognition},
  pages={17830--17839},
  year={2023}
}

@inproceedings{qiu2025gatedattention,
  title={Gated Attention for Large Language Models: Non-linearity, Sparsity, and Attention-Sink-Free},
  author={Qiu, Zihan and Wang, Zekun and Zheng, Bo and Huang, Zeyu and Wen, Kaiyue and Yang, Songlin and Men, Rui and Yu, Le and Huang, Fei and Huang, Suozhi and Liu, Dayiheng and Zhou, Jingren and Lin, Junyang},
  booktitle={Advances in Neural Information Processing Systems},
  year={2025},
}

@inproceedings{huang2017VR,
  title={6-DOF VR videos with a single 360-camera},
  author={Huang, Jingwei and Chen, Zhili and Ceylan, Duygu and Jin, Hailin},
  booktitle={2017 IEEE Virtual Reality (VR)},
  pages={37--44},
  year={2017},
  organization={IEEE}
}

@inproceedings{sun2019horizonnet,
  title={Horizonnet: Learning room layout with 1d representation and pano stretch data augmentation},
  author={Sun, Cheng and Hsiao, Chi-Wei and Sun, Min and Chen, Hwann-Tzong},
  booktitle={Proceedings of the IEEE/CVF Conference on Computer Vision and Pattern Recognition},
  pages={1047--1056},
  year={2019}
}

@inproceedings{hudson2019gqa,
  title={Gqa: A new dataset for real-world visual reasoning and compositional question answering},
  author={Hudson, Drew A and Manning, Christopher D},
  booktitle={Proceedings of the IEEE/CVF conference on computer vision and pattern recognition},
  pages={6700--6709},
  year={2019}
}

@inproceedings{xu2015image_captioning,
  title={Show, attend and tell: Neural image caption generation with visual attention},
  author={Xu, Kelvin and Ba, Jimmy and Kiros, Ryan and Cho, Kyunghyun and Courville, Aaron and Salakhudinov, Ruslan and Zemel, Rich and Bengio, Yoshua},
  booktitle={International conference on machine learning},
  pages={2048--2057},
  year={2015},
  organization={PMLR}
}

@inproceedings{antol2015vqa,
  title={Vqa: Visual question answering},
  author={Antol, Stanislaw and Agrawal, Aishwarya and Lu, Jiasen and Mitchell, Margaret and Batra, Dhruv and Zitnick, C Lawrence and Parikh, Devi},
  booktitle={Proceedings of the IEEE international conference on computer vision},
  pages={2425--2433},
  year={2015}
}

@article{wang2025multi,
  title={Multi-view Panoramic Image Style Transfer with Multi-scale Attention and Global Sharing},
  author={Wang, Weiyu and Qing, Chunmei and Tan, Junpeng and Xu, XiangMin},
  journal={ACM Transactions on Multimedia Computing, Communications and Applications},
  year={2025},
  publisher={ACM New York, NY}
}

@article{liu2023visual_instruction_tuning,
  title={Visual instruction tuning},
  author={Liu, Haotian and Li, Chunyuan and Wu, Qingyang and Lee, Yong Jae},
  journal={Advances in neural information processing systems},
  volume={36},
  pages={34892--34916},
  year={2023}
}

@inproceedings{li2023blip_2,
  title={Blip-2: Bootstrapping language-image pre-training with frozen image encoders and large language models},
  author={Li, Junnan and Li, Dongxu and Savarese, Silvio and Hoi, Steven},
  booktitle={International conference on machine learning},
  pages={19730--19742},
  year={2023},
  organization={PMLR}
}

@article{hu2022lora,
  title={Lora: Low-rank adaptation of large language models.},
  author={Hu, Edward J and Shen, Yelong and Wallis, Phillip and Allen-Zhu, Zeyuan and Li, Yuanzhi and Wang, Shean and Wang, Lu and Chen, Weizhu and others},
  journal={ICLR},
  volume={1},
  number={2},
  pages={3},
  year={2022}
}

@article{jiang2021unifuse,
  title={Unifuse: Unidirectional fusion for 360 panorama depth estimation},
  author={Jiang, Hualie and Sheng, Zhe and Zhu, Siyu and Dong, Zilong and Huang, Rui},
  journal={IEEE Robotics and Automation Letters},
  volume={6},
  number={2},
  pages={1519--1526},
  year={2021},
  publisher={IEEE}
}

@inproceedings{zhang2014panocontext,
  title={Panocontext: A whole-room 3d context model for panoramic scene understanding},
  author={Zhang, Yinda and Song, Shuran and Tan, Ping and Xiao, Jianxiong},
  booktitle={European conference on computer vision},
  pages={668--686},
  year={2014},
  organization={Springer}
}

@inproceedings{zhang2021deeppanocontext,
  title={Deeppanocontext: Panoramic 3d scene understanding with holistic scene context graph and relation-based optimization},
  author={Zhang, Cheng and Cui, Zhaopeng and Chen, Cai and Liu, Shuaicheng and Zeng, Bing and Bao, Hujun and Zhang, Yinda},
  booktitle={Proceedings of the IEEE/CVF International Conference on Computer Vision},
  pages={12632--12641},
  year={2021}
}

@inproceedings{philion2020lss,
    title={Lift, Splat, Shoot: Encoding Images From Arbitrary Camera Rigs by Implicitly Unprojecting to 3D},
    author={Jonah Philion and Sanja Fidler},
    booktitle={Proceedings of the European Conference on Computer Vision},
    year={2020},
}

@inproceedings{coors2018spherenet,
  title={Spherenet: Learning spherical representations for detection and classification in omnidirectional images},
  author={Coors, Benjamin and Condurache, Alexandru Paul and Geiger, Andreas},
  booktitle={Proceedings of the European conference on computer vision (ECCV)},
  pages={518--533},
  year={2018}
}

@inproceedings{shen2022panoformer,
  title={PanoFormer: panorama transformer for indoor $360^o$ depth estimation},
  author={Shen, Zhijie and Lin, Chunyu and Liao, Kang and Nie, Lang and Zheng, Zishuo and Zhao, Yao},
  booktitle={European Conference on Computer Vision},
  pages={195--211},
  year={2022},
  organization={Springer}
}

@inproceedings{zheng2024ops,
  title={Open panoramic segmentation},
  author={Zheng, Junwei and Liu, Ruiping and Chen, Yufan and Peng, Kunyu and Wu, Chengzhi and Yang, Kailun and Zhang, Jiaming and Stiefelhagen, Rainer},
  booktitle={European Conference on Computer Vision},
  pages={164--182},
  year={2024},
  organization={Springer}
}

@article{hu2024deformable,
  title={Deformable mamba for wide field of view segmentation},
  author={Hu, Jie and Zheng, Junwei and Wei, Jiale and Zhang, Jiaming and Stiefelhagen, Rainer},
  journal={arXiv preprint arXiv:2411.16481},
  year={2024}
}

@inproceedings{zheng2025spr,
  title={Scene-agnostic pose regression for visual localization},
  author={Zheng, Junwei and Liu, Ruiping and Chen, Yufan and Chen, Zhenfang and Yang, Kailun and Zhang, Jiaming and Stiefelhagen, Rainer},
  booktitle={Proceedings of the IEEE/CVF Conference on Computer Vision and Pattern Recognition},
  pages={27092--27102},
  year={2025}
}

@article{kaplan2020scaling_law,
  title={Scaling laws for neural language models},
  author={Kaplan, Jared and McCandlish, Sam and Henighan, Tom and Brown, Tom B and Chess, Benjamin and Child, Rewon and Gray, Scott and Radford, Alec and Wu, Jeffrey and Amodei, Dario},
  journal={arXiv preprint arXiv:2001.08361},
  year={2020}
}

@article{hoffmann2022training_scaling_law,
  title={Training compute-optimal large language models},
  author={Hoffmann, Jordan and Borgeaud, Sebastian and Mensch, Arthur and Buchatskaya, Elena and Cai, Trevor and Rutherford, Eliza and Casas, Diego de Las and Hendricks, Lisa Anne and Welbl, Johannes and Clark, Aidan and others},
  journal={arXiv preprint arXiv:2203.15556},
  year={2022}
}

@article{child2019window_attention,
  title={Generating long sequences with sparse transformers},
  author={Child, Rewon and Gray, Scott and Radford, Alec and Sutskever, Ilya},
  journal={arXiv preprint arXiv:1904.10509},
  year={2019}
}

@inproceedings{cao2024oass_blendpass,
  title={Occlusion-Aware Seamless Segmentation},
  author={Yihong Cao and Jiaming Zhang and Hao Shi and Kunyu Peng and Yuhongxuan Zhang and Hui Zhang and Rainer Stiefelhagen and Kailun Yang},
  booktitle={European Conference on Computer Vision (ECCV)},
  year={2024}
}

@inproceedings{kim2025vru_accident,
  title={VRU-Accident: A Vision-Language Benchmark for Video Question Answering and Dense Captioning for Accident Scene Understanding},
  author={Kim, Younggun and Abdelrahman, Ahmed S and Abdel-Aty, Mohamed},
  booktitle={Proceedings of the IEEE/CVF International Conference on Computer Vision},
  pages={761--771},
  year={2025}
}

@inproceedings{wang2024deepaccident,
  title={Deepaccident: A motion and accident prediction benchmark for v2x autonomous driving},
  author={Wang, Tianqi and Kim, Sukmin and Wenxuan, Ji and Xie, Enze and Ge, Chongjian and Chen, Junsong and Li, Zhenguo and Luo, Ping},
  booktitle={Proceedings of the AAAI Conference on Artificial Intelligence},
  volume={38},
  number={6},
  pages={5599--5606},
  year={2024}
}

@inproceedings{yang2021_wildPASS,
title={Capturing Omni-Range Context for Omnidirectional Segmentation},
author={Yang, Kailun and Zhang, Jiaming and Rei{\ss}, Simon and Hu, Xinxin and Stiefelhagen, Rainer},
booktitle={2021 IEEE/CVF Conference on Computer Vision and Pattern Recognition (CVPR)},
year={2021}
}

@inproceedings{gurari2018vizwiz,
  title={Vizwiz grand challenge: Answering visual questions from blind people},
  author={Gurari, Danna and Li, Qing and Stangl, Abigale J and Guo, Anhong and Lin, Chi and Grauman, Kristen and Luo, Jiebo and Bigham, Jeffrey P},
  booktitle={Proceedings of the IEEE conference on computer vision and pattern recognition},
  pages={3608--3617},
  year={2018}
}

@article{ying2025mmwalk,
  title={mmWalk: Towards Multi-modal Multi-view Walking Assistance},
  author={Ying, Kedi and Liu, Ruiping and Chen, Chongyan and Tao, Mingzhe and Shi, Hao and Yang, Kailun and Zhang, Jiaming and Stiefelhagen, Rainer},
  journal={arXiv preprint arXiv:2510.11520},
  year={2025}
}

@article{tian2025nuscenes_spatialqa,
  title={{NuScenes-spatialQA}: A spatial understanding and reasoning benchmark for vision-language models in autonomous driving},
  author={Tian, Kexin and Mao, Jingrui and Zhang, Yunlong and Jiang, Jiwan and Zhou, Yang and Tu, Zhengzhong},
  journal={arXiv preprint arXiv:2504.03164},
  year={2025}
}

@inproceedings{qian2024nuscenes_qa,
  title={{NuScenes-QA}: A multi-modal visual question answering benchmark for autonomous driving scenario},
  author={Qian, Tianwen and Chen, Jingjing and Zhuo, Linhai and Jiao, Yang and Jiang, Yu-Gang},
  booktitle={Proceedings of the AAAI Conference on Artificial Intelligence},
  volume={38},
  number={5},
  pages={4542--4550},
  year={2024}
}

@inproceedings{park2025nuplanqa,
  title={{NuPlanQA}: A large-scale dataset and benchmark for multi-view driving scene understanding in multi-modal large language models},
  author={Park, Sung-Yeon and Cui, Can and Ma, Yunsheng and Moradipari, Ahmadreza and Gupta, Rohit and Han, Kyungtae and Wang, Ziran},
  booktitle={ICCV},
  year={2025}
}

@inproceedings{wang2025omnidrive,
  title={{OmniDrive}: A Holistic Vision-Language Dataset for Autonomous Driving with Counterfactual Reasoning},
  author={Shihao Wang and Zhiding Yu and Xiaohui Jiang and Shiyi Lan and Min Shi and Nadine Chang and Jan Kautz and Ying Li and Jose M. Alvarez},
  booktitle={CVPR},
  year={2025}
}

@inproceedings{sima2024drivelm,
  title={Drivelm: Driving with graph visual question answering},
  author={Sima, Chonghao and Renz, Katrin and Chitta, Kashyap and Chen, Li and Zhang, Hanxue and Xie, Chengen and Bei{\ss}wenger, Jens and Luo, Ping and Geiger, Andreas and Li, Hongyang},
  booktitle={European conference on computer vision},
  pages={256--274},
  year={2024},
  organization={Springer}
}

@inproceedings{wei2024onebev,
  title={OneBEV: Using One Panoramic Image for Bird, Aos-Eye-View Semantic Mapping},
  author={Wei, Jiale and Zheng, Junwei and Liu, Ruiping and Hu, Jie and Zhang, Jiaming and Stiefelhagen, Rainer},
  booktitle={Proceedings of the Asian Conference on Computer Vision},
  pages={583--596},
  year={2024}
}

@article{li2025DA2,
  title={$\text{DA}^2$: Depth Anything in Any Direction},
  author={Li, Haodong and Zheng, Wangguangdong and He, Jing and Liu, Yuhao and Lin, Xin and Yang, Xin and Chen, Ying-Cong and Guo, Chunchao},
  journal={arXiv preprint arXiv:2509.26618},
  year={2025}
}

@article{li2024llms,
  title={Llms-as-judges: a comprehensive survey on llm-based evaluation methods},
  author={Li, Haitao and Dong, Qian and Chen, Junjie and Su, Huixue and Zhou, Yujia and Ai, Qingyao and Ye, Ziyi and Liu, Yiqun},
  journal={arXiv preprint arXiv:2412.05579},
  year={2024}
}

@article{li2025preference,
  title={Preference leakage: A contamination problem in llm-as-a-judge},
  author={Li, Dawei and Sun, Renliang and Huang, Yue and Zhong, Ming and Jiang, Bohan and Han, Jiawei and Zhang, Xiangliang and Wang, Wei and Liu, Huan},
  journal={arXiv preprint arXiv:2502.01534},
  year={2025}
}

@inproceedings{zhang2022bending_deformable,
  title={Bending reality: Distortion-aware transformers for adapting to panoramic semantic segmentation},
  author={Zhang, Jiaming and Yang, Kailun and Ma, Chaoxiang and Rei{\ss}, Simon and Peng, Kunyu and Stiefelhagen, Rainer},
  booktitle={Proceedings of the IEEE/CVF conference on computer vision and pattern recognition},
  pages={16917--16927},
  year={2022}
}

@ARTICLE{zhang2024deformable_panorama,
  author={Zhang, Jiaming and Yang, Kailun and Shi, Hao and Reiß, Simon and Peng, Kunyu and Ma, Chaoxiang and Fu, Haodong and Torr, Philip H. S. and Wang, Kaiwei and Stiefelhagen, Rainer},
  journal={IEEE Transactions on Pattern Analysis and Machine Intelligence}, 
  title={Behind Every Domain There is a Shift: Adapting Distortion-Aware Vision Transformers for Panoramic Semantic Segmentation}, 
  year={2024},
  volume={46},
  number={12},
  pages={8549-8567},
  doi={10.1109/TPAMI.2024.3408642}}

@article{team2023gemini,
  title={Gemini: a family of highly capable multimodal models},
  author={Team, Gemini and Anil, Rohan and Borgeaud, Sebastian and Wu, Yonghui and Alayrac, Jean-Baptiste and Yu, Jiahui and Soricut, Radu and Schalkwyk, Johan and Dai, Andrew M and Hauth, Anja and others},
  journal={arXiv preprint arXiv:2312.11805},
  year={2023}
}

@article{hong2025glm,
  title={Glm-4.1 v-thinking: Towards versatile multimodal reasoning with scalable reinforcement learning},
  author={Hong, Wenyi and Yu, Wenmeng and Gu, Xiaotao and Wang, Guo and Gan, Guobing and Tang, Haomiao and Cheng, Jiale and Qi, Ji and Ji, Junhui and Pan, Lihang and others},
  journal={arXiv e-prints},
  pages={arXiv--2507},
  year={2025}
}

@inproceedings{chen2024internvl,
  title={Internvl: Scaling up vision foundation models and aligning for generic visual-linguistic tasks},
  author={Chen, Zhe and Wu, Jiannan and Wang, Wenhai and Su, Weijie and Chen, Guo and Xing, Sen and Zhong, Muyan and Zhang, Qinglong and Zhu, Xizhou and Lu, Lewei and others},
  booktitle={Proceedings of the IEEE/CVF conference on computer vision and pattern recognition},
  pages={24185--24198},
  year={2024}
}

@article{bai2023qwen_vl,
  title={Qwen-vl: A frontier large vision-language model with versatile abilities},
  author={Bai, Jinze and Bai, Shuai and Yang, Shusheng and Wang, Shijie and Tan, Sinan and Wang, Peng and Lin, Junyang and Zhou, Chang and Zhou, Jingren},
  journal={arXiv preprint arXiv:2308.12966},
  year={2023}
}

@article{lu2024deepseek_vl,
  title={Deepseek-vl: towards real-world vision-language understanding},
  author={Lu, Haoyu and Liu, Wen and Zhang, Bo and Wang, Bingxuan and Dong, Kai and Liu, Bo and Sun, Jingxiang and Ren, Tongzheng and Li, Zhuoshu and Yang, Hao and others},
  journal={arXiv preprint arXiv:2403.05525},
  year={2024}
}

@article{Qwen2_5_vl2025,
  title={Qwen2.5-VL Technical Report},
  author={Bai, Shuai and Chen, Keqin and Liu, Xuejing and Wang, Jialin and Ge, Wenbin and Song, Sibo and Dang, Kai and Wang, Peng and Wang, Shijie and Tang, Jun and Zhong, Humen and Zhu, Yuanzhi and Yang, Mingkun and Li, Zhaohai and Wan, Jianqiang and Wang, Pengfei and Ding, Wei and Fu, Zheren and Xu, Yiheng and Ye, Jiabo and Zhang, Xi and Xie, Tianbao and Cheng, Zesen and Zhang, Hang and Yang, Zhibo and Xu, Haiyang and Lin, Junyang},
  journal={arXiv preprint arXiv:2502.13923},
  year={2025}
}

@misc{glm2024chatglm,
      title={ChatGLM: A Family of Large Language Models from GLM-130B to GLM-4 All Tools}, 
      author={Team GLM and Aohan Zeng and Bin Xu and Bowen Wang and Chenhui Zhang and Da Yin and Diego Rojas and Guanyu Feng and Hanlin Zhao and Hanyu Lai and Hao Yu and Hongning Wang and Jiadai Sun and Jiajie Zhang and Jiale Cheng and Jiayi Gui and Jie Tang and Jing Zhang and Juanzi Li and Lei Zhao and Lindong Wu and Lucen Zhong and Mingdao Liu and Minlie Huang and Peng Zhang and Qinkai Zheng and Rui Lu and Shuaiqi Duan and Shudan Zhang and Shulin Cao and Shuxun Yang and Weng Lam Tam and Wenyi Zhao and Xiao Liu and Xiao Xia and Xiaohan Zhang and Xiaotao Gu and Xin Lv and Xinghan Liu and Xinyi Liu and Xinyue Yang and Xixuan Song and Xunkai Zhang and Yifan An and Yifan Xu and Yilin Niu and Yuantao Yang and Yueyan Li and Yushi Bai and Yuxiao Dong and Zehan Qi and Zhaoyu Wang and Zhen Yang and Zhengxiao Du and Zhenyu Hou and Zihan Wang},
      year={2024},
      eprint={2406.12793},
      archivePrefix={arXiv},
      primaryClass={id='cs.CL' full_name='Computation and Language' is_active=True alt_name='cmp-lg' in_archive='cs' is_general=False description='Covers natural language processing. Roughly includes material in ACM Subject Class I.2.7. Note that work on artificial languages (programming languages, logics, formal systems) that does not explicitly address natural-language issues broadly construed (natural-language processing, computational linguistics, speech, text retrieval, etc.) is not appropriate for this area.'}
}

@article{xu2025chatbev,
  title={ChatBEV: A Visual Language Model that Understands BEV Maps},
  author={Xu, Qingyao and Chen, Siheng and Chen, Guang and Wang, Yanfeng and Zhang, Ya},
  journal={arXiv preprint arXiv:2503.13938},
  year={2025}
}

@misc{team2024internvl2,
  title={Internvl2: Better than the best—expanding performance boundaries of open-source multimodal models with the progressive scaling strategy},
  author={Team, OpenGVLab},
  year={2024},
  publisher={Accessed}
}

@article{zhu2025internvl3,
  title={Internvl3: Exploring advanced training and test-time recipes for open-source multimodal models},
  author={Zhu, Jinguo and Wang, Weiyun and Chen, Zhe and Liu, Zhaoyang and Ye, Shenglong and Gu, Lixin and Tian, Hao and Duan, Yuchen and Su, Weijie and Shao, Jie and others},
  journal={arXiv preprint arXiv:2504.10479},
  year={2025}
}

@article{li2024llava_onevision,
  title={Llava-onevision: Easy visual task transfer},
  author={Li, Bo and Zhang, Yuanhan and Guo, Dong and Zhang, Renrui and Li, Feng and Zhang, Hao and Zhang, Kaichen and Zhang, Peiyuan and Li, Yanwei and Liu, Ziwei and others},
  journal={arXiv preprint arXiv:2408.03326},
  year={2024}
}

@article{liu2023llava1.5,
  title={Improved baselines with visual instruction tuning},
  author={Liu, Haotian and Li, Chunyuan and Li, Yuheng and Lee, Yong Jae},
  journal={arXiv preprint arXiv:2310.03744},
  year={2023}
}
}

\clearpage
\setcounter{page}{1}
\maketitlesupplementary

\appendix


\section{Sparse Attention}
\subsection{Preliminary of Simplified Sparse Attention}
To verify the effectiveness of sparse attention with an efficient indexing mechanism in panoramic tasks, we implement a simple yet effective sparse attention, called Simplified Sparse Attention (SSA). Similar to PSA, SSA employs a lightweight indexer to select the Top-K key tokens. However, SSA simplifies the similarity scoring by removing the position-aware gating network. Specifically, the indexer computes a score matrix $\mathbf{I} \in \mathbb{R}^{L \times L}$. The score $I_{t,s}$ between a query token $t$ and a key token $s$ is calculated by
\vskip -0.5em
\begin{align}
   I_{t,s} = \sum_{j=1}^{H^I} w_{t,j} \cdot \operatorname{ReLU}(\mathbf{q}_{I,(t,j)}^\top \mathbf{k}_{I,(s,j)}),
\label{eq:ssa_score}
\end{align}
where $H^I$ is the number of selector heads. For each head $j$, the query vector $\mathbf{q}_{I,(t,j)} \in \mathbb{R}^{d_I}$ and key vector $\mathbf{k}_{I,(s,j)} \in \mathbb{R}^{d_I}$ are obtained via linear projections of the input states $\mathbf{h}_t^{(l-1)}$ and $\mathbf{h}_s^{(l-1)}$. Differ from the gate mechanism in PSA, the weighting term $w_{t,j}$ is derived solely from the query token, i.e.,
\begin{align}
   \mathbf{w}_t^{(l)} = \mathbf{h}_t^{(l-1)} \mathbf{W}_{w}^{(l)},
\end{align}
where $\mathbf{W}_{w}^{(l)} \in \mathbb{R}^{d \times H^I}$, and $w_{t,j}$ is the $j$-th scalar component of $\mathbf{w}_t^{(l)}$, representing the importance of head $j$ for the current query $t$. The ReLU activation ensures sparsity in the dot-product similarity. Based on the score matrix $\mathbf{I}$, we select the indices of the Top-K relevant keys for each query $t$, denoted as $\mathcal{S}_t = \{s \mid I_{t,s} \in \text{Top-K}(\mathbf{I}_{t,:}) \}$. The subsequent attention computation remains consistent with the sparse paradigm:
\begin{align}
   \mathbf{o}_t^{(l)}= \text{Attention}\left( \mathbf{q}_t^{(l)}, \mathbf{K}_{\mathcal{S}_t}^{(l)}, \mathbf{V}_{\mathcal{S}_t}^{(l)} \right).
\end{align}

Finally, the output of the SSA is produced by aggregating the token outputs and applying a linear projection:
\begin{align}
   \text{SSA}(\mathbf{h}^{(l-1)}) &= \text{Stack}(\mathbf{o}_1^{(l)}, \dots, \mathbf{o}_{L}^{(l)}) \mathbf{W}_{O,\text{sparse}}^{(l)}.
\end{align}

\subsection{Discussion on Proposed PSA}
\begin{figure*}[t]
    \centering
    \includegraphics[width=1.0\linewidth]{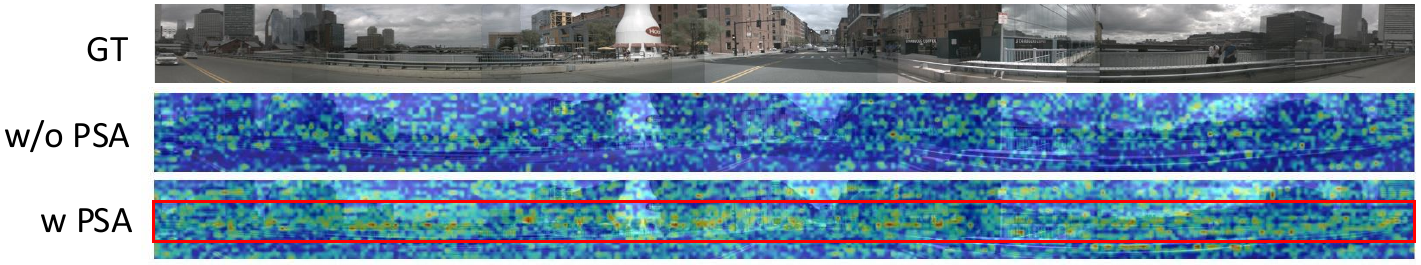}
    \caption{Attention visualization of our proposed PSA. PSA filters uninformative regions like ``sky'' and distant backgrounds, while concentrating on the critical navigation areas (e.g., the road and surrounding vehicles) across the 360-degree panoramic view.}
    \label{fig:rebuttal_example}
    \vspace{-1em}
\end{figure*}

We highlight three key points of our PSA: (1) 
Unlike pixel-level Deformable CNNs, PSA utilizes \textit{patch-wise} attention. This is specifically designed for high-resolution panoramas to capture long-range semantic dependencies (\eg, wrap-around continuity) that pixel-based methods miss. (2) 
We introduce a gating mechanism (Eq.~\ref{eq:gate}) as a semantic noise filter to prune vast redundant regions (\eg, sky). This integration of gating and sparsity is supported by recent findings in efficient LLMs~\cite{qiu2025gatedattention} for handling massive token sequences. (3) 
As shown in Fig.~\ref{fig:rebuttal_example}, the effectiveness of PSA is clear: while the baseline scatters attention noisily, PSA concentrates activation strictly along the semantic horizon (buildings, roads marked in \textcolor{red}{red box}), suppressing uninformative backgrounds.

Furthermore, to test the compatibility of our proposed PSA, we integrated it with the LLaVA-Next-7B and conducted experiments on the PanoVQA-mini dataset. The results are shown in Table~\ref{tab:compatibility_of_psa}, the addition of PSA enhances performance across all subsets, confirming PSA's robust plug-and-play capability.

\begin{table}[t]
    \centering
    \caption{compatibility of PSA on LLaVA-Next-7B}
    \label{tab:compatibility_of_psa}
    \resizebox{\columnwidth}{!}{
    \setlength{\tabcolsep}{1.0mm}
        \begin{tabular}{lcccc}
            \toprule
            \textbf{Model} & \textbf{Avg(N)} & \textbf{Avg(O)} & \textbf{Avg(D)} & \textbf{Avg} \\
            \midrule
            LLaVA-Next-7B       & 15.36 & 34.14 & 26.25 & 24.59 \\
            LLaVA-Next-7B + PSA & 16.05 & 33.57 & 28.53 & 25.63 \\
            \bottomrule
        \end{tabular}
    }
\end{table}


\section{PanoVQA Benchmark}

\subsection{Data Collection}
\noindent \textbf{PanoVQA-N (NuScenes)} 
As a cornerstone for real-world autonomous driving perception, the NuScenes dataset~\cite{caesar2020nuscenes} comprises 1,000 scenes collected in Boston and Singapore. It provides high-quality data from a sensor suite including 6 cameras, LiDAR, and Radar, offering a complete $360^\circ$ field of view. We utilize the multi-view camera images to construct panoramic inputs, serving as the primary source for training the model's fundamental perception capabilities, such as object detection, and spatial relationships.

\noindent \textbf{PanoVQA-O (BlendPASS).} 
To evaluate the performance of existing VLMs, particularly in occlusion scenarios, we incorporate the BlendPASS dataset. Utilizing a native $360^\circ$ camera, this dataset captures a seamless, full-surround Field of View (FoV). BlendPASS is characterized by its high-fidelity rendering and diverse visual scenarios, which are often underrepresented in standard real-world collections. By including this data, we significantly improve the model's robustness against varying lighting conditions and environmental contexts, ensuring consistent performance across different visual domains.

\noindent  \textbf{PanoVQA-D (DeepAccident).}
Standard datasets like NuScenes rarely contain collision data, limiting a model's ability to reason about safety-critical situations. DeepAccident~\cite{wang2024deepaccident} is a large-scale synthetic dataset explicitly designed for accident prediction and causal reasoning. It covers diverse accident types and complex interaction scenarios. We leverage this dataset to construct accident-related QA pairs, requiring the model not only to perceive the environment but also to predict potential hazards and explain the underlying causes of accidents, which is crucial for safety-aware autonomous driving.

\subsection{Generating Details}
\label{sec:gen_details}
Our dataset generation pipeline integrates three distinct autonomous driving sources, including NuScenes, BlendPASS, and DeepAccident, to construct a large-scale panoramic VQA benchmark. The process comprises three sequential stages: (1) geometry-based panoramic synthesis, (2) structured annotation preprocessing, and (3) automated QA generation with quality assurance.



\subsubsection{Panoramic Image Synthesis}
For the NuScenes and DeepAccident datasets, we synthesize panoramic images using a geometry-based stitching approach following by OneBEV~\cite{wei2024onebev}. We model the environment using a viewing sphere centered at the optical origin $O$. Each of the six camera views, denoted as $\mathcal{I}_i$ for $i \in \{1, \dots, 6\}$, is treated as a 2D image plane tangent to this sphere at a specific point $M_i$, aligned with the camera's extrinsic optical axis.

The pixel mapping is computed via ray casting. For each target pixel $P_{j,k}$ in the panoramic coordinate system, we cast a 3D ray $\vec{l}_{j,k}$ outward from the origin $O$. The source pixel correspondence is established by calculating where this ray intersects the camera planes:
\begin{align}
    \mathbf{N}_{j,k}^{(i)} &= \operatorname{Intersect}(\vec{l}_{j,k}, \mathcal{I}_i), \\
    \mathcal{P}(P_{j,k}) &= \mathcal{I}_{i^*}(\text{Proj}(\mathbf{N}_{j,k}^{(i^*)})),
\end{align}
where $\mathcal{P}(P_{j,k})$ denotes the assigned pixel value (e.g., RGB color) in the final panorama $\mathcal{P}$ at spatial coordinate $(j,k)$. $\mathbf{N}_{j,k}^{(i)}$ represents the 3D intersection point of the ray with the $i$-th camera plane, and $\text{Proj}(\cdot)$ projects this 3D point back into the 2D pixel coordinates of that specific camera to fetch the source value. 

Since the cameras' fields of view naturally overlap, a single ray might hit multiple valid camera planes. To prevent blurry ghosting artifacts caused by pixel blending, we resolve these overlaps using a strict deterministic priority rule: $i^* = \min \{i \mid \mathbf{N}_{j,k}^{(i)} \in \text{Valid}(\mathcal{I}_i)\}$. This guarantees spatial consistency by exclusively selecting the valid camera with the highest priority (the lowest index $i^*$). Note that the BlendPASS dataset provides native panoramic imagery, thereby bypassing this entire stitching phase.

\subsubsection{Structured Annotation Processing}
To standardize object properties across heterogeneous sources, we use a quadruple to denote each object in the scene. Each object of interest is encoded as a tuple $\mathcal{Q} = (c, \theta, d, v/s)$, representing category, direction, distance, and visibility/speed, respectively.

\paragraph{PanoVQA-N (NuScenes Baseline).} 
We derive spatial attributes from 3D bounding boxes. The direction $\theta$ is the angular displacement relative to the ego-vehicle's heading vector $\vec{v}_{\text{ego}}$, and visibility $v$ is original meta informations. 

\paragraph{PanoVQA-O (BlendPASS Occlusion).} 
Focusing on occlusion handling, we employ the DA2~\cite{li2025DA2} for pixel-level depth estimation. For an object represented by a polygonal region $\mathcal{R}$, the distance is computed as the regional average:
\begin{align}
    \bar{D} = \frac{1}{|\mathcal{R}|} \sum_{(u,v) \in \mathcal{R}} D_{(u,v)},
\end{align}
where $D(u,v)$ is the estimated depth at pixel $(u,v)$. Occlusion relationships are determined by comparing the geometric intersection and $y$-axis ordering of object polygons.

\paragraph{PanoVQA-D (DeepAccident).} 
To incorporate temporal dynamics from static imagery, we extend the quadruple format to include speed $s$. The tuple becomes $\mathcal{Q}_{D} = (c, \theta, d, s)$, where ground-truth speed data is used to compensate for the ambiguity of motion inference in single-frame inputs.

\begin{figure*}[tbp!]
  \centering
  \includegraphics[width=0.95\linewidth]{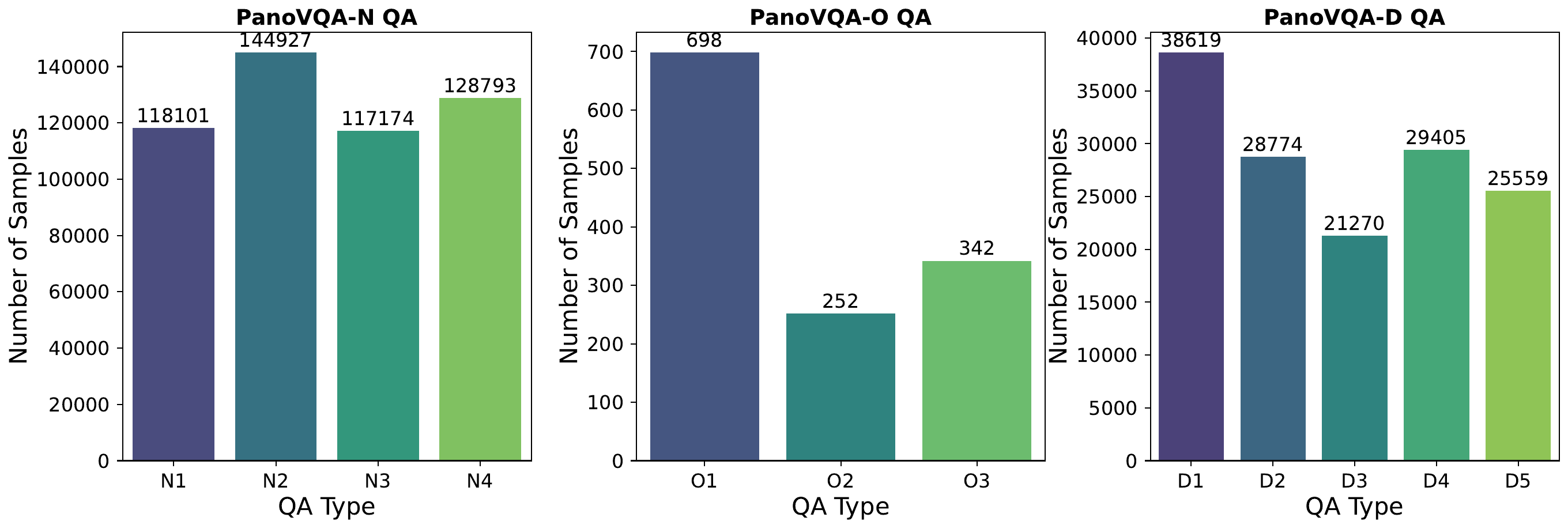}
  \vskip -0.5em
  \caption{Distribution of QA samples in PanoVQA. The dataset features disparate scales to mimic real-world distributions: PanoVQA-N (left) offers a large-scale foundation ($\sim$510k), PanoVQA-O (middle) introduces a challenging long-tailed setting ($\sim$1.3k), and PanoVQA-D (right) covers safety-critical events ($\sim$143k).}
  \vskip -1.5em
  \label{fig:data_stat}
\end{figure*}

\begin{figure}[t!]
\centering
\begin{tcolorbox}[
  width=\linewidth,
  colback=gray!5,
  colframe=black!70,
  arc=2mm,
  boxrule=0.4pt,
  left=2pt, right=2pt, top=2pt, bottom=2pt
]
\begin{lstlisting}[
  language=bash,                 % 设置语言为 bash (适合命令行)
  basicstyle=\ttfamily\scriptsize, % 使用等宽字体，字号设为小号以防溢出
  keywordstyle=\color{blue},     % 关键字颜色
  commentstyle=\color{codegreen},% 注释颜色
  stringstyle=\color{codepurple},% 字符串颜色
  breaklines=true,               % 自动换行
  breakindent=16pt,              % 换行后的缩进
  columns=flexible,              % 字符间距灵活
  showstringspaces=false,        % 不显示字符串中的空格符号
  frame=none                     % 关闭 lstlisting 自带的边框 (因为外面已经有 tcolorbox 了)
]
qwenvl/train/train_qwen.py 
  --model_name_or_path "Qwen2.5-VL-7B-Instruct" 
  --run_name "qwen2.5-vl-finetune" 
  --deepspeed "scripts/zero3.json" 
  --dataset_use "NuScenes_train%60,DeepAccident_train%90" 
  --data_flatten True 
  --tune_mm_vision True 
  --tune_mm_mlp True 
  --tune_mm_llm True 
  --tune_mm_adaptorformer False 
  --bf16 True 
  --output_dir "./output/train_whole/7B(vit,mlp,llm)_baseline" 
  --num_train_epochs 1.0 
  --per_device_train_batch_size 8 
  --per_device_eval_batch_size 16 
  --gradient_accumulation_steps 1 
  --max_pixels 50176 
  --min_pixels 784 
  --save_strategy "steps" 
  --save_steps 100 
  --save_total_limit 1 
  --learning_rate 5e-6 
  --weight_decay 0 
  --warmup_ratio 0.03 
  --max_grad_norm 1 
  --lr_scheduler_type "cosine" 
  --logging_steps 1 
  --model_max_length 8192 
  --gradient_checkpointing True 
  --dataloader_num_workers 8 
  --report_to "wandb"
\end{lstlisting}
\end{tcolorbox}
\caption{Hyperparameters for Qwen2.5-VL Baseline.}
\label{fig:qwen_finetune_baseline}
\end{figure}

\subsubsection{Automated QA Generation and Statistics}
We employ \texttt{GPT-5-mini} to generate QA pairs, incorporating structured quadruples $\mathcal{Q}$ into the template shown in Fig.~\ref{fig:prompt_for_generating_panovqa}. To optimize the trade-off between reasoning depth and efficiency, we apply adaptive reasoning effort settings: \texttt{minimal} for standard driving scenes (PanoVQA-N) and \texttt{low} for the more challenging scenarios in PanoVQA-O and PanoVQA-D.

\textbf{Quality Assurance.} The pipeline includes a two-stage validation: (1) \textit{Automated Filtering} to verify semantic alignment between the generated QA and source annotations; (2) \textit{Human Evaluation} to assess correctness and fluency.

\textbf{Statistics.} The final PanoVQA benchmark comprises a total of 538,635 training samples and 115,279 validation samples. The dataset features an average question length of 19.07 words and an answer length of 42.4 words, ensuring sufficient complexity for comprehensive evaluation across normal, occluded, and accident-prone scenarios.

\textbf{Panoramic QA.} To ensure the model processes the full $360^{\circ}$ context (i.e., panoramic scene), we emphasize panoramic scene in three ways. (1) The system prompt explicitly include "panoramic scene" (see Figure~\ref{fig:prompt_for_generating_panovqa}) to force generate panorama-related qa pairs during generation. (2) The questions utilize keywords like ``\textit{panoramic scene}'' and ``\textit{whole scene}'' to trigger global reasoning. Our quadruple tuple includes a direction field that compels the model to explicitly model spatial relationships.

\subsection{Data Statistics}
As illustrated in Figure~\ref{fig:data_stat}, PanoVQA spans diverse scales while maintaining internal balance. \textbf{PanoVQA-N} provides a massive foundation ($>$500k) with samples evenly distributed across four QA types. Similarly, \textbf{PanoVQA-D} ($\sim$143k) maintains a relatively uniform distribution across its five accident categories. In the \textbf{PanoVQA-O} subset ($\sim$1.3k), the data remains distributed across its types without extreme skew. This internal balance prevents the model from overfitting to specific question patterns, while the global scale disparity effectively tests robust generalization.

\subsection{Human Evaluation}
\label{sec:human_eval}
To assess annotation fidelity, we implemented a dual-verification protocol on the \textit{PanoVQA-mini} subset involving five human experts. We partitioned the evaluation samples equally between human reviewers and GPT-4o; the consistent scoring distributions observed across both groups attest to the dataset's reliability.
Quantitatively, as detailed in Table~\ref{tab:quality_score_opt}, 96\% of the samples were classified as ``Valid'' or ``High Quality.'' Furthermore, the high degree of alignment—with a deviation of less than 0.5\% between GPT-4o and human evaluations—confirms that our automated pipeline serves as a robust proxy for human judgment.

\section{Implementation Details}

\subsection{Training Script}
To ensure the reproducibility of our experiments, we present the detailed training command and hyperparameter configuration in Fig.~\ref{fig:qwen_finetune_baseline}. We utilize \texttt{Qwen2.5-VL-7B-Instruct} as our base foundation model. We use supervised fine-tuning and unfreezing all components, including the visual encoder, MLP-based merger, and LLM.

\subsection{Evaluation Prompt}
Fig.~\ref{fig:eval_gpt_prompt} illustrates the prompt utilized to score the model inference on a scale of $1{\sim}5$. After obtaining the raw responses from the evaluator, our post-processing script automatically parses the numerical ratings and aggregates them to assess model performance from multiple perspectives. Specifically, it computes normalized scores for each item based on our pre-defined categories, respectively. Finally, the system derives a comprehensive overall score, reported with a precision of two decimal places, to facilitate accurate quantitative comparisons.

\begin{table}[t!]
    \centering
    \caption{Comparison of multi-view (6 cameras) and panoramic (1 camera) modeling across Normal (N), Occluded (O), and Accidental (D) scenes.}
    \label{tab:performance_comparison_6camer_pano}
    \vskip -0.5em
    \resizebox{\columnwidth}{!}{
    \setlength{\tabcolsep}{1.2mm}
        \begin{tabular}{lccccc}
            \toprule
            \multirow{2}{*}{\textbf{Model}} & \multirow{2}{*}{\textbf{Input Strategy}} & \multicolumn{4}{c}{\textbf{Metrics} ($\uparrow$)} \\
            \cmidrule(lr){3-6}
            & & Avg(N) & Avg(O) & Avg(D) & \textbf{Avg} \\
            \midrule
            \multirow{2}{*}{Qwen2.5-VL-3B}
            & Multi-view (6 cam) & 16.82 & \textbf{32.40} & \textbf{34.92} & \textbf{28.26} \\
            & Uni-view (${2\times}3$ grid)   & 15.69 & 31.27 & 30.39 & 25.71 \\
            & Panoramic (1 pano) & \textbf{16.86} & 30.44 & 31.25 & 26.25 \\
            \midrule
            \multirow{3}{*}{Qwen2.5-VL-3B-SFT}
            & Multi-view (6 cam) & 26.33 & 39.88 & \textbf{54.45} & 40.22 \\
            & Uni-view (${2\times}3$ grid) & 26.04 & 40.52 & 50.95 & 40.04 \\
            & Panoramic (1 pano) & \textbf{29.68} & \textbf{40.98} & 51.08 & \textbf{41.42} \\
            \bottomrule
        \end{tabular}
    }
\end{table}


\begin{figure*}[h!]
\centering
\begin{tcolorbox}[
  width=\textwidth,              
  colback=gray!5,
  colframe=black!70,
  arc=2mm,
  boxrule=0.4pt,
  left=1pt, right=1pt, top=1pt, bottom=5pt  
]
\begin{lstlisting}[
  basicstyle=\rmfamily,
  breaklines=true,
  breakindent=0pt,
  columns=fullflexible
]

SYSTEM_MESSAGE = ``Generate 15 QA pairs from one panoramic scene JSON file from perspectives from users. Each scene includes object annotations (category, attributes, and relative direction). Generate Question Answer pairs for each panoramic scene, which should cover all 4 QA Types. (At least 1 pair for each type) The Question Answer pairs must follow the instructions and rules below, taking the given sample as an example.''

PROMPT = ``
Rules: 
1. Analyze the panoramic scene annotations, focusing on:
    - Use a quadruple tuple (category, direction, distance, visibility) to describe an object (e.g., `a fully visible pedestrian in the back right around 9 meters'). Use vague numbers to express distances, without decimal points.
    - Object attributes and spatial relationships (visibility, distance, and direction relative to the ego car).
2. Only describe clear information in the images, do not fabricate or invent in the answers. 
3. Base all answers only on what is actually visible in the provided json data. Do not make assumptions or invent details. 
4. All positions and absolute coordinates must be described in a directional manner. (Describe exact direction such as `front left', `back right', `front', etc.)
5. Visibility Encoding:
    1: Low visibility (0-40%)
    2: Medium visibility (40-60%)
    3: High visibility (60-80%)
    4: Fully visible (80-100%)
6. The question can be slightly modified to produce different answers. For multi-item answers, maintain the order relevant to the question (e.g., nearest to farthest). One question should correspond to one answer.
7. All responses should be written expressions in natural language, avoid using symbols or brackets.

Instructions:
Fully consider following levels to generate questions and multiple answers:
1. Short Level QA: QA pairs that query the basic information in the json file or single panoramic image, the answer can be completely verified by the ground truth. 
2. Long Level QA: QA pairs that contain multiple objects, with attributions and their relationships in concern, the answer stems mainly from the combined ground truth feature information. The questions should be short and rough, while the answers should be detailed and comprehensive. The answer can be partially verified.

QA Types:
-Type N1- Global scene understanding (overall scene description, including objects, visibility)
-Type N2- Object and attribute identification (attributes of objects present)
-Type N3- Relationship identification (object-object spatial relationships)
-Type N4- Location description (where objects are located in relation to ego car, ego-object spatial relationships)

JSON file:
{json_data}
''
\end{lstlisting}
\end{tcolorbox}
\caption{Prompt used in PanoVQA generation. Using PanoVQA-N as an example.}
\label{fig:prompt_for_generating_panovqa}
\end{figure*}

\begin{figure*}[h!]
\centering
\begin{tcolorbox}[
  width=\textwidth,
  colback=gray!5,
  colframe=black!70,
  arc=2mm,
  boxrule=0.4pt,
  left=1pt, right=1pt, top=1pt, bottom=1pt
]
\begin{lstlisting}[
  basicstyle=\rmfamily,
  breaklines=true,
  breakindent=0pt,
  columns=fullflexible
]
messages = [
    {
        ``role'': ``system'',
        ``content'':
            ``You are an intelligent evaluator designed to evaluate the correctness and similarity of generative outputs for question-answer pairs. ''
            ``Your task is to compare the model prediction answer with the correct answer and determine if they match in meaning. Here's the scoring criteria:\n\n''
            "### Scoring Criteria:\n''
            ``5 = Perfect match or Correct in meaning\n''
            ``4 = Key information correct, minor flaws\n''
            ``3 = Partially correct\n''
            ``2 = Mostly wrong answer for key query, but some relevance\n''
            ``1 = Completely wrong or nonsense sentences\n\n''
            ``Your response must ONLY be the integer score (e.g., 4). DO NOT include any text or explanation.''
    },
    {
        ``role'': ``user'',
        ``content'':
            f``Question: {question}\n''
            f``Correct Answer: {gt_answer}\n''
            f``Predicted Answer: {pred_answer}\n\n''
            ``Please provide a score from 1 to 5 based on how well the predicted answer matches the correct answer.''
    }
]
\end{lstlisting}
\end{tcolorbox}
\caption{Prompt used in evaluation.}
\label{fig:eval_gpt_prompt}
\end{figure*}

\section{More Experiment Analysis} 
\subsection{Qualitative Results}
To systematically evaluate the effectiveness of our panoramic approach, we compare three distinct visual input modalities for the VLM. As summarized in Table~\ref{tab:performance_comparison_6camer_pano}, the experimental settings are defined as follows:

\begin{compactitem}
    \item \textbf{Multi-view (6-cam):} The 6 surrounding camera images are input as independent sequences of visual token. We utilize text prompts to identify each view (e.g., ``Front: ${<}$image1${>}$, Front-Left: ${<}$image2${>}$$\dots$''), relying on the LLM to mentally reconstruct the spatial topology.
    \item \textbf{Uni-view (${2\times}3$ grid):} The 6 images are spatially concatenated into a single $2 {\times} 3$ grid image before feeding into the vision encoder. This preserves the original resolution without an extremely unbalanced aspect ratio.
    \item \textbf{Panoramic (1-pano):} Our proposed method, where images are cylindrically projected and stitched. Note that this process involves \textit{vertical cropping} to remove distortion and blank areas, resulting in a reduction of total pixels compared to the raw 6-camera input.
\end{compactitem}

\noindent \textbf{Analysis.} 
The quantitative results in Table~\ref{tab:performance_comparison_6camer_pano} reveal a compelling finding. While the Multi-view approach preserves the highest image fidelity and vertical Field of View (FoV), it forces the model to process fragmented visual information. In contrast, despite the inevitable resolution loss and vertical cropping inherent in the stitching process, the \textbf{Panoramic} input achieves superior performance after Supervised Fine-Tuning (SFT), reaching an overall score of \textbf{41.42} compared to \textbf{40.22} for the Multi-view baseline.

This performance gain is particularly evident in \textit{PanoVQA-N} (29.68 vs. 26.33) and \textit{PanoVQA-O} (40.98 vs. 39.88) scenarios. We attribute this to the \textbf{seamless spatial context} provided by the panorama. The continuous $360^\circ$ view eliminates the cognitive burden of stitching disjointed images, allowing the model to better understand spatial relationships and object continuity across camera boundaries. This demonstrates that for holistic scene understanding in autonomous driving, spatial continuity is often more critical than maximizing pixel count.


\begin{figure}[t]
  \centering
  \includegraphics[width=1.0\linewidth]{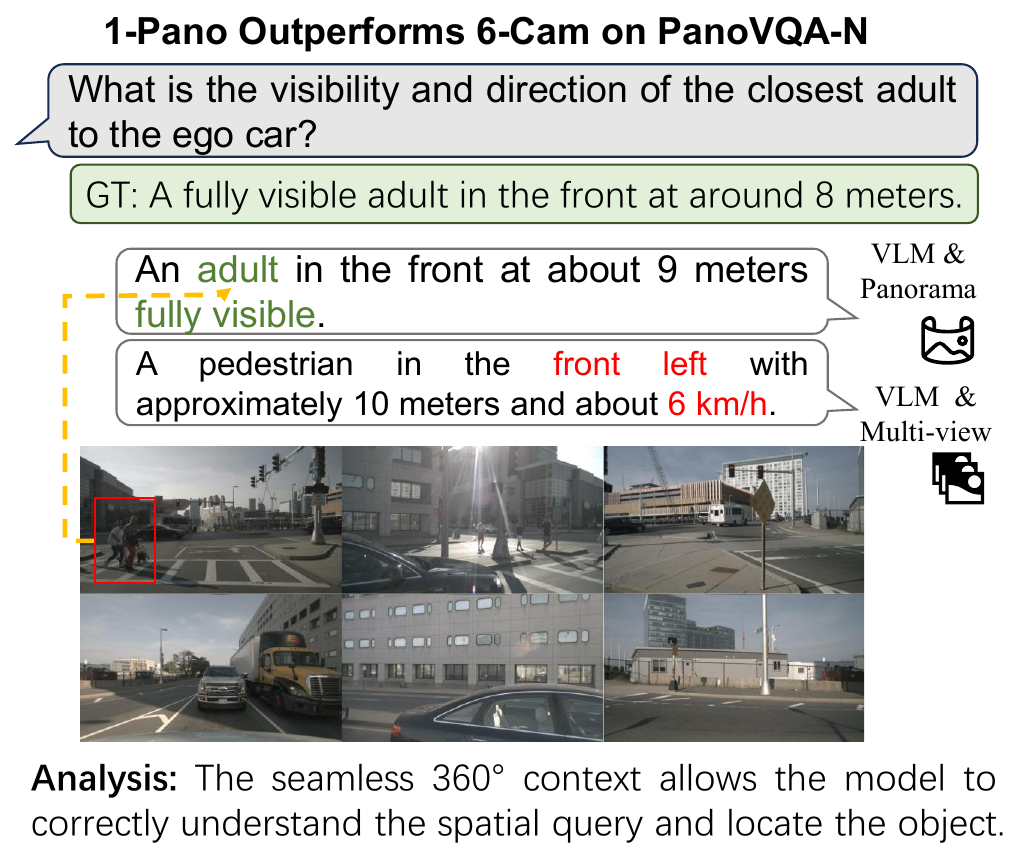}
  \vskip -0.5em
  \caption{Qualitative comparison on \textbf{PanoVQA-N}. The panoramic model (1-Pano) correctly identifies the spatial location (``front'') and visibility of the pedestrian, whereas the multi-view model (6-Cam) hallucinates a ``front left'' direction due to fragmented spatial context.}
  \vskip -1.5em
  \label{fig:exp1_panovqa_n}
\end{figure}

\subsection{More Results on PanoVQA-mini}
We additionally evaluate several advanced proprietary models on PanoVQA-mini, including Gemini-3-Pro, GPT-5, and Claude-Sonnet-4.5. As shown in~\ref{tab:proprietary_results}, their performance indicates that panoramic understanding remains a challenge for generalist models, likely due to the scarcity of panoramic data in their pre-training corpora.
\begin{table}[t]
\centering
\small
\caption{Proprietary results on PanoVQA-mini.}
\label{tab:proprietary_results}
\begin{tabular}{lcccc}
\toprule
\textbf{Model} & \textbf{Avg(N)} & \textbf{Avg(O)} & \textbf{Avg(D)} & \textbf{Avg} \\
\midrule
Gemini-3-Pro      & 18.04 & 20.86 & 37.32 & 26.78 \\
GPT-5             & 22.02 & 40.55 & 51.64 & 38.99 \\
Claude-Sonnet-4.5 & 19.29 & 41.73 & 52.76 & 38.84 \\
\bottomrule
\end{tabular}
\end{table}

\subsection{More Examples}
\noindent \textbf{Reasoning in Normal Scenarios.} 
Fig.~\ref{fig:exp1_panovqa_n} illustrates a perception task from PanoVQA-N, where the model is asked to identify the visibility and location of the closest adult. The 1-Pano model accurately locates the adult in the ``front'' at approximately 9 meters and correctly identifies them as ``fully visible.'' In contrast, the 6-Cam model hallucinates the direction as ``front left'' and provides extraneous speed information not present in the visual cues. This discrepancy highlights that the seamless $360^\circ$ context of the panoramic input aids the model in establishing a more precise ego-centric coordinate system, whereas the fragmented multi-view input may introduce spatial ambiguities across camera overlaps.

\begin{figure}[t]
  \centering
  \includegraphics[width=1.0\linewidth]{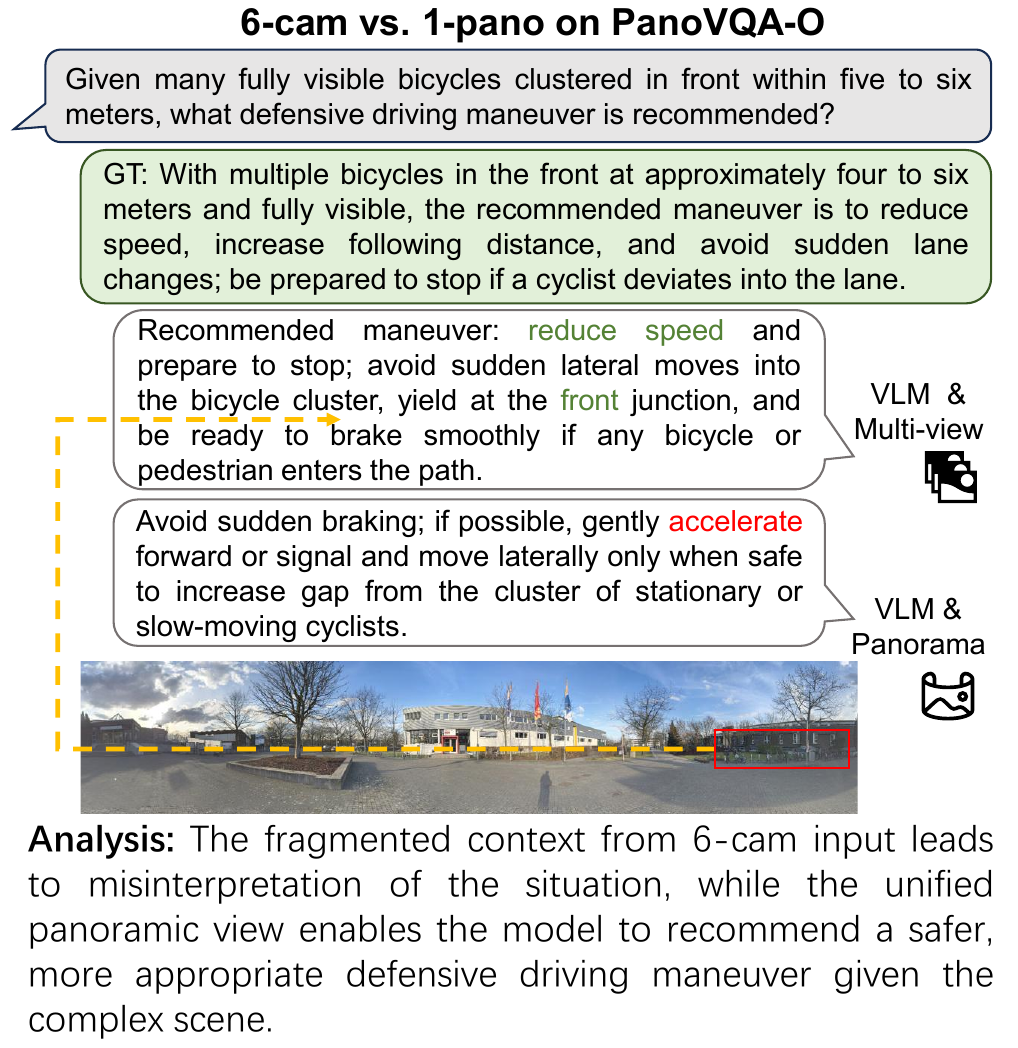}
  \vskip -0.5em
  \caption{Qualitative comparison on \textbf{PanoVQA-O}. Facing a cluster of bicycles, the 1-Pano model benefits from the unified view to propose a coherent defensive maneuver, demonstrating the importance of seamless context for planning.}
  \vskip -1.5em
  \label{fig:exp1_panovqa_o}
\end{figure}

\noindent \textbf{Reasoning under Occlusion.} 
In the complex scenario shown in Fig.~\ref{fig:exp1_panovqa_o} (PanoVQA-O), the vehicle encounters a cluster of bicycles. The prompt requires a defensive driving maneuver. The 6-Cam model's reasoning appears fragmented, potentially due to the bicycle cluster spanning across camera boundaries. Conversely, the 1-Pano model leverages the unified view to holistically assess the density of the cluster, recommending a contextually safer maneuver. This suggests that spatial continuity is critical for planning tasks where objects of interest (e.g., crowds, large vehicles) span multiple viewing angles.

\begin{figure}[t]
  \centering
  \includegraphics[width=1.0\linewidth]{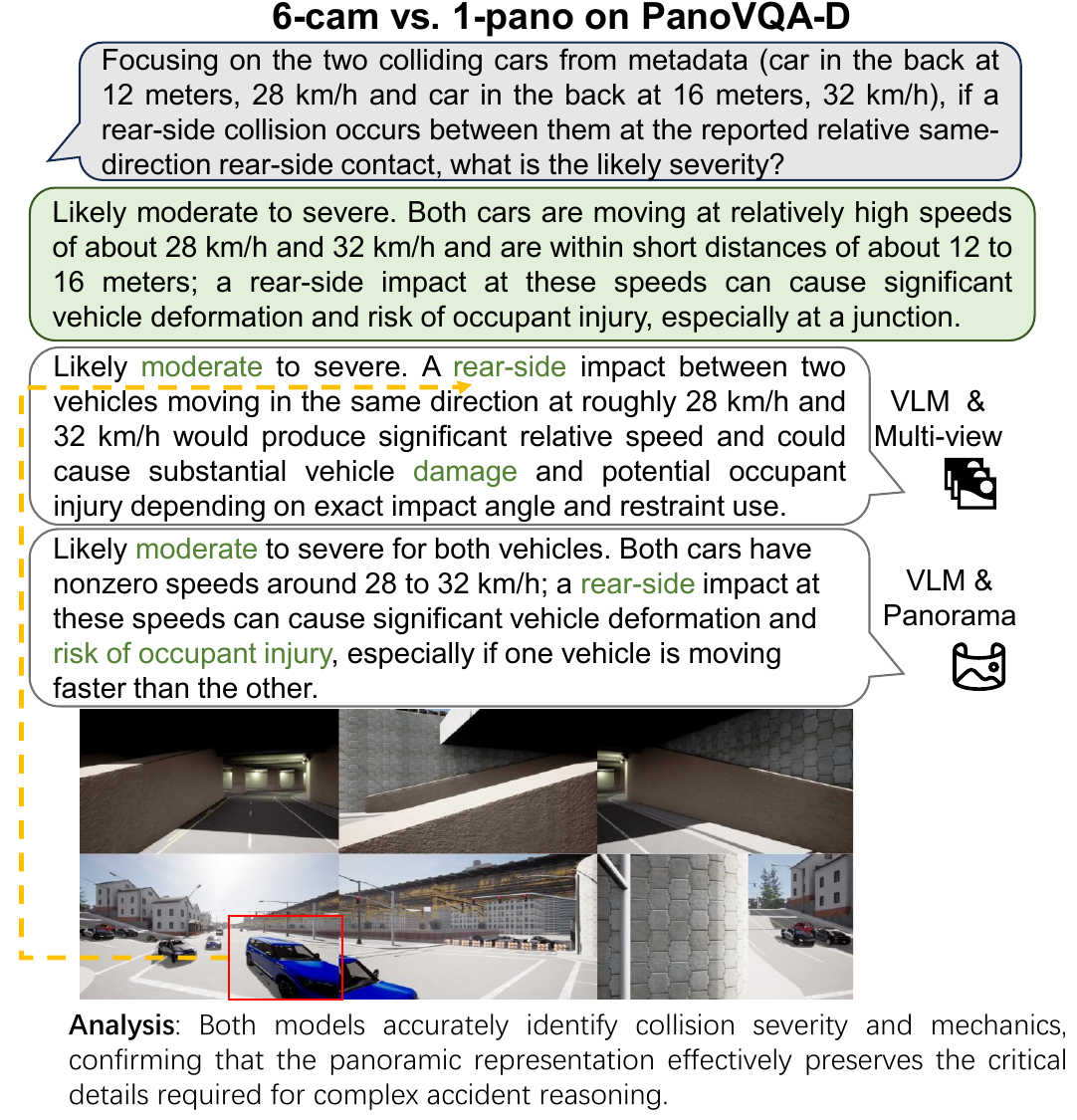}
  \vskip -0.5em
  \caption{Qualitative comparison on \textbf{PanoVQA-D}. Both models accurately predict the severity and type of collision. This confirms that the panoramic representation retains critical visual details necessary for complex accident reasoning.}
  \vskip -1.5em
  \label{fig:exp1_panovqa_d}
\end{figure}

\noindent \textbf{Reasoning in Safety-Critical scenarios.} 
Fig.~\ref{fig:exp1_panovqa_d} presents a rear-end collision scenario from PanoVQA-D. Interestingly, both the 6-Cam and 1-Pano models successfully identify the collision mechanics (``rear-side impact'') and accurately estimate the severity as ``moderate to severe.'' This result is significant as it validates that our panoramic stitching and cropping process—despite the theoretical loss of vertical field-of-view—preserves the essential fine-grained visual details required for high-level accident analysis.

\noindent \textbf{Conclusion.} 
The qualitative results corroborate our quantitative findings: while multi-view inputs preserve raw pixel fidelity, the \textbf{1-Pano representation offers superior spatial coherence}. This coherence proves advantageous in tasks requiring precise localization and holistic scene understanding, without compromising performance on semantic reasoning tasks like accident analysis.


\section{Limitation and Future Work}
\label{sec:limitation}
Despite the robust spatial reasoning demonstrated by our panoramic framework, the current geometric stitching process introduces inevitable limitations. To achieve a seamless projection, the method necessitates vertical cropping and induces boundary distortions, resulting in pixel-level information loss. This reduction in fidelity explains instances where raw multi-view inputs outperform the panoramic representation, especially in zero-shot settings or scenarios demanding high-resolution visual detail.

Future research will focus on enhancing visual fidelity and expanding into the temporal domain. We plan to explore more advanced image stitching techniques to reduce information loss and preserve better visual details compared to current fixed projection methods. Furthermore, acknowledging the dynamic nature of autonomous driving, we aim to extend our framework to process video inputs, enabling the model to understand movement and unfolding events rather than just static scenes.

\end{document}